\documentclass[11pt]{article}

\usepackage[final]{acl}

\usepackage{times}
\usepackage{latexsym}

\usepackage[T1]{fontenc}

\usepackage[utf8]{inputenc}

\usepackage{microtype}

\usepackage{inconsolata}

\usepackage{graphicx}
\usepackage{amsmath,amssymb}
\usepackage{booktabs}
\usepackage{multirow}
\usepackage{tabularx}
\usepackage{algorithm}
\usepackage{algpseudocode}
\usepackage{array}
\usepackage[most]{tcolorbox}
\usepackage{listings}
\usepackage{enumitem}
\tcbuselibrary{breakable,listings}

\lstdefinestyle{promptstyle}{
    basicstyle=\ttfamily\scriptsize,
    breaklines=true,
    breakatwhitespace=true,
    columns=fullflexible,
    keepspaces=true,
    showstringspaces=false
}

\newtcblisting{promptbox}[2][]{
    enhanced,
    breakable,
    listing only,
    colback=gray!5,
    colframe=gray!50,
    colbacktitle=gray!25,
    coltitle=black,
    boxrule=0.5pt,
    arc=2mm,
    left=1.2mm,
    right=1.2mm,
    top=0.8mm,
    bottom=0.8mm,
    title={#2},
    fonttitle=\bfseries\small,
    listing options={style=promptstyle},
    #1
}
\newcolumntype{Y}{>{\raggedright\arraybackslash}X}

\newcommand{\KL}{\operatorname{KL}}
\newcommand{\cival}[2]{#1 {\scriptsize #2}}

\title{When Quantization Preserves Accuracy but Not Evidence: Explanation-Aware Post-Training Quantization for Medical LLMs}

\author{
  \textbf{Yeji Kim},
  \textbf{Mi-Young Kim},
  \textbf{Randy Goebel}
  \\
  University of Alberta
  \\
  \small{
  \texttt{\{yeji7, miyoung2, rgoebel\}@ualberta.ca}
  }
}

\begin{document}
\maketitle
\begin{abstract}
Post-training quantization (PTQ) enables efficient deployment of large language models, and PTQ methods are usually optimized and evaluated with generic reconstruction, perplexity, or answer accuracy. But in explanation-critical domains, preserving only the final answer may be insufficient, since users may also inspect generated rationales to judge whether a prediction is trustworthy. We study this issue in medical multiple-choice question answering, where rationales should provide evidence that supports the selected answer.

We propose an explanation-aware objective for transformation-based PTQ. Our method builds an offline faithfulness cache from full-precision teacher rationales and uses it during optimization to preserve answer-supporting evidence tokens and evidence-conditioned answer behavior. We instantiate it on OSTQuant under W4A4KV4 quantization and evaluate four 7B--8B medical and instruction-tuned LLMs on MedExQA, MedExpQA, and ChallengeClinicalQA. While a same-calibration OSTQuant baseline preserves task accuracy, it can substantially weaken answer-supporting rationales. Our objective is to preserve the full-precision model's answer-supporting behavior rather than improve gold-label accuracy, and our method better preserves the full-precision model's answer behavior and rationale-to-answer support. These results suggest that PTQ for explanation-critical settings should evaluate preservation of answer-supporting evidence, not only answer accuracy. Code and evaluation scripts are available at https://github.com/dut0817/EAQuant.
\end{abstract}

\section{Introduction}

Post-training quantization (PTQ) reduces the memory and inference cost of large language models (LLMs) by lowering the precision of weights, activations, and KV-cache states without full retraining. Standard LLM PTQ methods aim to preserve full-precision behavior by optimizing against reconstruction error, perplexity, calibration losses, or downstream answer accuracy~\citep{frantar2022gptq,xiao2023smoothquant,lin2023awq,shao2024omniquant}. These metrics are appropriate for generic compression fidelity, but they do not directly test whether a quantized model preserves the explanation behavior that users may inspect and rely on.

\begin{figure}[t]
    \centering
    \includegraphics[width=\columnwidth]{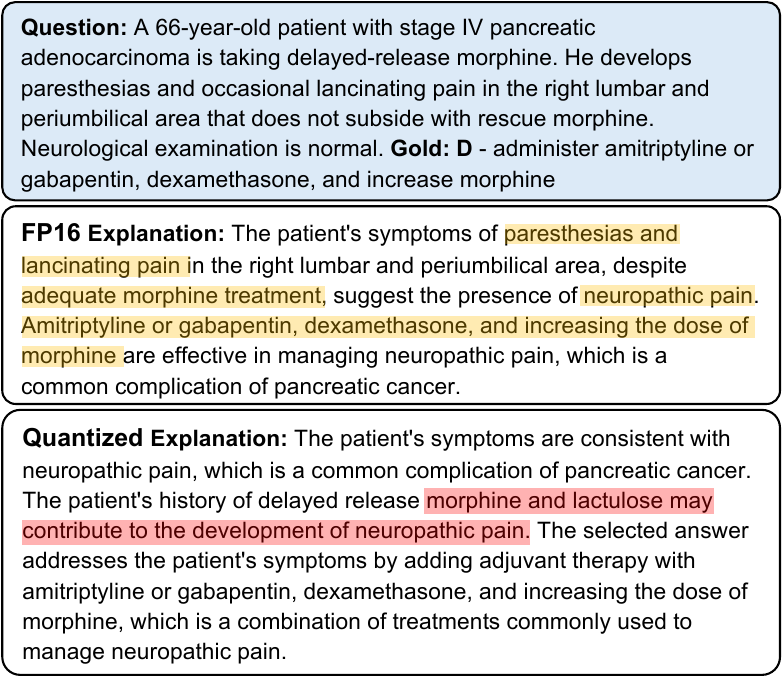}
    \caption{
    Motivating example of rationale drift after quantization.
    Yellow marks answer-supporting evidence from the full-precision reference, and red marks an unsupported claim introduced after quantization.
    }
    \label{fig:motivating_rationale_drift}
\end{figure}

The difference between preserving what a model answers and preserving how it justifies that answer is especially consequential in explanation-critical domains. For example, in medical QA, users may not only ask whether a model selected the correct option, but also \emph{whether the generated rationale cites the clinical evidence that supports that option}. Prior work distinguishes faithful explanations from plausible explanations and emphasizes that rationales should be meaningfully associated with model decisions, not merely sound convincing~\citep{jacovi-goldberg-2020-towards,wiegreffe-etal-2021-measuring,hase2020leakage}. Recent clinical-generation work similarly stresses preserving important clinical information and avoiding unsupported content~\citep{xie-etal-2024-doclens,oukelmoun2025detecting}. Therefore a quantized model can be problematic even when answer accuracy appears stable. The quantized model may preserve the answer while changing, omitting, or fabricating the answer-supporting evidence.

Figure~\ref{fig:motivating_rationale_drift} illustrates this failure mode. The full-precision and quantized models select the same answer, but the quantized rationale changes the clinical justification. This motivates our central question: \emph{when quantization preserves answers, does it also preserve the evidence that supports those answers?} A recent study of LLM self-explanations under quantization suggests that quantization can affect explanation quality and faithfulness across bit-widths and methods~\citep{wang2026can}, but to our knowledge no prior work directly modifies the PTQ objective to preserve explanation behavior. We extend this inquiry to medical multiple-choice QA  using MedExQA, MedExpQA, and ChallengeClinicalQA~\citep{kim-etal-2024-medexqa,alonso2024medexpqa,chen-etal-2025-benchmarking}.

We propose an explanation-aware objective for transformation-based PTQ. Our goal is to preserve the full-precision model's answer-supporting behavior under quantization, rather than to improve gold-label accuracy. To avoid directly reinforcing known teacher errors, we construct the cache only from calibration examples for which the full-precision teacher's prediction matches the gold answer. Our method is designed for PTQ pipelines with calibration-time transformation optimization. We instantiate it on OSTQuant~\citep{hu2025ostquant}, a strong transformation-based method for low-bit weight, activation, and KV-cache quantization. Our method keeps the calibration pool, deployed W4A4KV4 format, and fused inference path unchanged. During transformation optimization, it builds an offline faithfulness cache from full-precision teacher rationales and adds two extra preservation losses to the optimization objective. The cache selects answer-supporting evidence spans from the teacher rationale, such as phrases or clauses, that support the teacher's selected answer and become less answer-supporting after quantization. The added losses preserve both evidence-token behavior and evidence-conditioned answer behavior, so that the quantized model better retains the answer-supporting role of the full-precision rationale.

We evaluate four 7B--8B medical and instruction-tuned LLMs on MedExQA, MedExpQA, and ChallengeClinicalQA under W4A4KV4 quantization. The results show that the OSTQuant baseline can substantially weaken answer-rationale support even when task accuracy is preserved. In contrast, our method better preserves the full-precision model's answer choices and rationale–answer consistency.

Our contributions are as follows:
\begin{itemize}[noitemsep,topsep=0pt,leftmargin=*]
    \item We expose a gap between task accuracy and rationale preservation in medical QA PTQ, where low-bit quantization can preserve task accuracy while degrading answer-supporting rationales.
    \item We introduce a teacher-rationale cache and explanation-aware transformation losses that preserve answer-supporting evidence tokens and rationale-conditioned answer behavior during calibration.
    \item We evaluate W4A4KV4 PTQ with answer and evidence metrics, showing improved preservation of full-precision answer behavior and rationale support over a calibration-matched baseline.
\end{itemize}

\section{Related Work}
\label{sec:related_work}
\paragraph{LLM post-training quantization (PTQ)} 
PTQ reduces memory and inference cost by quantizing weights, activations, and KV-cache states after training. 
generative pre-trained quantization (GPTQ) uses approximate second-order information for weight quantization~\citep{frantar2022gptq}; SmoothQuant addresses the difficulty of quantizing activation outliers by rescaling channels and shifting part of the quantization burden from activations to weights~\citep{xiao2023smoothquant}; activation-aware weight quantization (AWQ) uses activation statistics to identify and protect salient weight channels~\citep{lin2023awq}. More recent differentiable or transformation-based methods optimize quantization parameters or equivalent transformations during calibration, including OmniQuant~\citep{shao2024omniquant}, SpinQuant~\citep{liu2025spinquant}, and OSTQuant~\citep{hu2025ostquant}. All these methods mainly target numerical fidelity, perplexity, or task accuracy. Here we study whether the resulting quantized model also preserves answer-supporting rationale behavior.

\paragraph{Medical explanations and behavior-aware compression}
Medical QA benchmarks increasingly include rationales or explanations beyond answer labels~\citep{kim-etal-2024-medexqa,alonso2024medexpqa,chen-etal-2025-benchmarking}. Prior work distinguishes plausible explanations from faithful ones and argues that rationales should be meaningfully coupled with model decisions~\citep{jacovi-goldberg-2020-towards,wiegreffe-etal-2021-measuring,hase2020leakage}. Clinical generation work also emphasizes preserving important clinical information and avoiding unsupported content~\citep{xie-etal-2024-doclens,oukelmoun2025detecting}. Recent studies show that quantization can affect LLM self-explanations~\citep{wang2026can}, and that PTQ objectives can be adapted to preserve behavior-critical properties such as safety alignment~\citep{wee2025alignment}. We extend this direction to medical rationale preservation under low-bit PTQ.

\section{Diagnostic Analysis}
\label{sec:diagnostic_analysis}
Before introducing our method, we examine which rationale tokens receive signal from the original OSTQuant token-level fidelity loss. For this diagnostic, we compute token-level Kullback--Leibler (KL) (FP vs. OSTQuant next-token distribution) on full-precision rationales generated by Llama-3-8B-Instruct on MedExpQA, and cluster tokens into numeric cues, answer/option tokens, clinical cues, and generic/discourse tokens. The full clustering procedure is described in Appendix~\ref{sec:diagnostic_details}.

Table~\ref{tab:loss_cue_alignment} shows that generic/discourse tokens account for 75.3\% of rationale tokens but receive 83.1\% of the KL signal, whereas clinical cues receive a smaller signal share (15.6\%) than their token share (22.2\%). Thus, the base fidelity objective is dominated by common discourse tokens rather than the sparse clinical cues that make a rationale answer-supporting. This mirrors the failure mode in Figure~\ref{fig:motivating_rationale_drift}: a quantized model may preserve a fluent-looking rationale and even the same answer while changing the clinical justification. We therefore add explanation-aware losses that concentrate preservation pressure on answer-supporting clinical evidence.

\begin{table}[t]
\centering
\small
\setlength{\tabcolsep}{5.0pt}
\renewcommand{\arraystretch}{0.95}
\begin{tabular}{lcc}
\toprule
Token group & Token \% & Signal \% \\
\midrule
Numeric cues       & 0.67  & 0.38  \\
Answer/option      & 1.80  & 0.99  \\
Clinical cues      & 22.19 & 15.56 \\
Generic/discourse  & 75.34 & 83.07 \\
\bottomrule
\end{tabular}
\caption{
Distribution of OSTQuant token-level fidelity signal in the diagnostic setting.
Token \% is the fraction of full-precision rationale tokens in each group.
Signal \% is the fraction of total token-level KL divergence assigned to each group.
}
\label{tab:loss_cue_alignment}
\end{table}

\section{Method}
\label{sec:method}
This section presents our explanation-aware objective for transformation-based PTQ. Our method targets PTQ pipelines that optimize transformation parameters during calibration and fuse them into the final low-bit model. We add calibration-time losses that preserve answer-supporting evidence tokens and evidence-conditioned answer behavior from full-precision teacher rationales. We instantiate the method on OSTQuant~\citep{hu2025ostquant}, a strong recent transformation-based PTQ baseline.

\begin{figure*}[t]
    \centering
    \includegraphics[width=\textwidth]{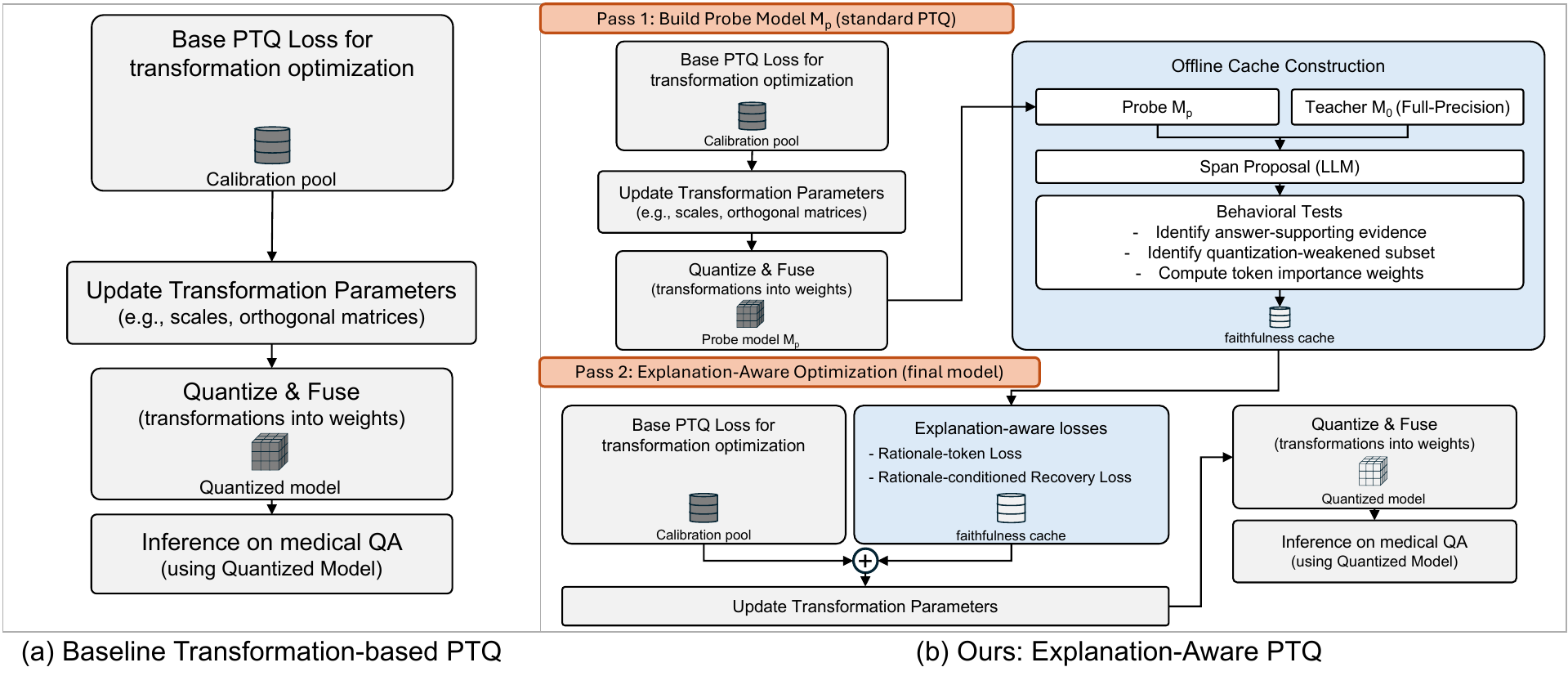}
    \caption{
    Overview of baseline transformation-based PTQ and our explanation-aware PTQ. Ours builds a quantized probe $M_p$ and an offline cache before cache-conditioned final optimization, while keeping the deployed quantized model format and inference cost unchanged.}
    \label{fig:overview}
\end{figure*}

\paragraph{PTQ optimization setup}
Figure~\ref{fig:overview} shows the intervention point. The OSTQuant baseline performs a single calibration-time transformation-optimization pass with the base PTQ objective, then quantizes and fuses the optimized transformations into the deployed model. Our method uses two calibration-time passes. In Pass 1, we run the same standard OSTQuant procedure on the same calibration pool to obtain a fixed quantized probe $M_p$. We use this probe only to construct the offline cache. In Pass 2, with the cache fixed, we optimize the final transformation parameters using the base PTQ loss together with the rationale-token and rationale-conditioned recovery losses. The deployed architecture, W4A4KV4 format, and inference path are identical to the OSTQuant baseline.

Within a transformation-optimization pass, let $M_{\phi}^{\mathrm{fp}}$ denote the transformed model before quantization, and let $M_{\phi}^{\mathrm{q}}$ denote the same transformed model under fake quantization, which simulates the target low-bit quantization during calibration.
Here, $\phi$ denotes the calibration-optimized transformation parameters.
OSTQuant optimizes $\phi$ by matching the next-token prediction distributions of these two paths on calibration text.
Let $u$ denote a calibration text sequence sampled from a calibration pool $\mathcal{D}_{\mathrm{cal}}$ (defined below).
We denote this base objective as
\begin{equation}
    \mathcal{L}_{\mathrm{OST}}
    =
    \mathbb{E}_{u\sim\mathcal{D}_{\mathrm{cal}}}
    \left[
    \ell_{\mathrm{KL\text{-}Top}}
    \left(
    M_{\phi}^{\mathrm{fp}},
    M_{\phi}^{\mathrm{q}};
    u
    \right)
    \right],
    \label{eq:ost_loss}
\end{equation}
Here, $\ell_{\text{KL-Top}}$ is the original OSTQuant KL loss computed over the top-K next-token probabilities of the quantization-disabled path $M^{\text{fp}}_\phi$.
It matches the fake-quantized path to the quantization-disabled transformed path. Our method keeps $\mathcal{L}_{\mathrm{OST}}$ and adds explanation-preserving losses during the second optimization pass.

Both OSTQuant and our method use the same non-test medical calibration pool $\mathcal{D}_{\mathrm{cal}}$. This controls for calibration-data effects, which can affect post-training compression performance~\citep{williams2024impact}. Let $x_i=(q_i,\mathcal{O}_i,y_i)$ be a medical multiple-choice example, where $q_i$ is the question, $\mathcal{O}_i=\{o_i^1,\ldots,o_i^{K_i}\}$ is the answer-option set with $K_i$ options, and $y_i\in\{1,\ldots,K_i\}$ is the gold option index.
We distinguish two full-precision objects. $M_0$ is the original full-precision teacher used only to build the fixed offline cache. $M_{\phi}^{\mathrm{fp}}$ is the quantization-disabled path of the transformed model during OSTQuant optimization.

\paragraph{Answer-option scoring}
To decide whether a rationale or evidence span supports an answer, we use deterministic forced-choice option scoring~\citep{holtzman2021surface,robinson2023leveraginglargelanguagemodels} rather than free-form answer parsing, which can be sensitive to decoding and answer-format choices~\citep{kumar-2022-answer,biderman2024lessons,tsvilodub2024predictions}. 
For a model $M$, option text
$o_i^k=(o_{i,1}^k,\ldots,o_{i,L_{i,k}}^k)$, where $L_{i,k}$ is the number of tokens in option $o_i^k$, and prompt $u$, we define
\begin{equation}
    a_{M,i}(k\mid u)
    =
    \frac{1}{L_{i,k}}
    \sum_{t=1}^{L_{i,k}}
    \log p_M(o_{i,t}^k\mid u,o_{i,<t}^k).
    \label{eq:option_score}
\end{equation}
The predicted option is
$\hat{y}_{M,i}(u)=\arg\max_{k} a_{M,i}(k\mid u)$.
When a distribution over options is needed, we apply a softmax over option scores and denote it by $\pi_{M,i}(\cdot\mid u)$. For target option $y_i$, the answer margin is
\begin{equation}
    m_{M,i}(u;y_i)
    =
    a_{M,i}(y_i\mid u)
    -
    \max_{k\neq y_i}a_{M,i}(k\mid u).
    \label{eq:answer_margin}
\end{equation}

\subsection{Faithfulness Cache Construction}
\label{sec:faithfulness_cache}
After obtaining the fixed same-calibration quantized probe $M_p$ in Pass 1, we build an offline faithfulness cache $\mathcal{A}$ from the shared non-test calibration pool. Its purpose is to identify which parts of a full-precision teacher rationale should receive additional preservation pressure under quantization. We use faithfulness cache as an internal term. Operationally, it stores evidence that supports the full-precision teacher's selected answer rather than establishing clinical factuality or causal explanation faithfulness.

For each calibration example, the full-precision teacher $M_0$ selects an answer using Eq.~\ref{eq:option_score} and then generates an answer-conditioned rationale $r_i$~\citep{hasan-etal-2026-reason2decide}. We keep only examples where the teacher prediction matches the gold option index $y_i$. 
Note that this filter is a reliability check, not a supervised answer loss. The gold label is used only to discard cache entries where the teacher itself is wrong. This prevents the cache from preserving rationales for teacher mistakes.

We use a prompted Qwen2.5-72B-Instruct model~\citep{qwen2.5} to generate high-recall candidate spans from the teacher rationale. For each candidate span $e$, we measure how strongly it supports the teacher answer using forced-choice option margins. We compare the teacher-answer margin both when $e$ is provided as evidence and when $e$ is removed from the rationale. A span is retained as teacher-answer-supporting evidence if it provides strong support as defined in Appendix~\ref{sec:cache_algorithm_appendix} for the teacher answer under these behavioral tests~\citep{deyoung2020eraser}.
We then repeat the same scoring with a same-calibration quantized probe $M_p$ and define $\mathcal{Q}_i\subseteq\mathcal{F}_i$ as the spans whose answer-supporting effect is weakened after quantization. To avoid trivially derivable answers, direct answer-label cues are removed before evidence extraction. The prompts used for cache construction are provided in Appendix~\ref{sec:prompts}. 

Each retained cache entry is
\[
\xi_i=(x_i,r_i,\mathcal{F}_i,\mathcal{Q}_i,\omega_i)\in\mathcal{A}.
\]
Here, $\mathcal{F}_i$ is the set of teacher-answer-supporting evidence spans, $\mathcal{Q}_i\subseteq\mathcal{F}_i$ is the subset whose answer-supporting effect is weakened by the quantized probe, and $\omega_i$ is a normalized token-level weight vector supported on tokens covered by $\mathcal{Q}_i$. We use $\mathcal{Q}_i$ for the rationale-token loss (focusing on tokens most affected by quantization) and $\mathcal{F}_i$ for the rationale-conditioned recovery loss (preserving the answer-supporting behavior of all teacher-answer-supporting evidence). The full cache construction details are provided in Appendix~\ref{sec:cache_algorithm_appendix}.

\subsection{Explanation-preserving Transformation Optimization}
\label{sec:explanation_optimization}
During the second transformation optimization pass, the base loss $\mathcal{L}_{\mathrm{OST}}$ is computed on the full calibration pool.
As previously noted, for cache entries $\xi_i=(x_i,r_i,\mathcal{F}_i,\mathcal{Q}_i,\omega_i)\sim\mathcal{A}$, we add two explanation-preserving losses. Both losses compare the fake-quantized path $M_{\phi}^{\mathrm{q}}$ against the quantization-disabled path $M_{\phi}^{\mathrm{fp}}$. They are not supervised answer losses. They preserve full-precision behavior on selected rationale evidence.

\paragraph{Rationale-token loss}
Let $c_i$ be the rationale-generation prompt, and let
$d_{i,t}$ be the next-token KL divergence between
$M_{\phi}^{\mathrm{fp}}$ and $M_{\phi}^{\mathrm{q}}$ at prefix
$(c_i,r_{i,<t})$:
\[
d_{i,t}
=
\KL\!\left(
p_{M_{\phi}^{\mathrm{fp}}}(\cdot\mid c_i,r_{i,<t})
\Vert
p_{M_{\phi}^{\mathrm{q}}}(\cdot\mid c_i,r_{i,<t})
\right).
\]
The rationale-token loss is
\begin{equation}
    \mathcal{L}_{\mathrm{rat}}
    =
    \mathbb{E}_{\xi_i\sim\mathcal{A}}
    \left[
    \sum_{t=1}^{T_i}
    \omega_{i,t} d_{i,t}
    \right].
    \label{eq:rationale_token_loss}
\end{equation}
This loss focuses token-level preservation on evidence that is both teacher-answer-supporting and quantization-affected, rather than imitating the entire rationale surface form.

\paragraph{Rationale-conditioned recovery loss}
Token preservation alone does not ensure that the same evidence induces the same answer preference after quantization.
For each teacher-answer-supporting evidence span $e\in\mathcal{F}_i$, we build a recovery prompt $u_i(e)$. 
Let
$h_{i,e}$ be the KL divergence between the option distributions of
$M_{\phi}^{\mathrm{fp}}$ and $M_{\phi}^{\mathrm{q}}$ under this prompt:
\[
h_{i,e}
=
\KL\!\left(
\pi_{M_{\phi}^{\mathrm{fp}},i}(\cdot\mid u_i(e))
\Vert
\pi_{M_{\phi}^{\mathrm{q}},i}(\cdot\mid u_i(e))
\right).
\]
The recovery loss is
\begin{equation}
    \mathcal{L}_{\mathrm{rec}}
    =
    \mathbb{E}_{\xi_i\sim\mathcal{A}}
    \left[
    \frac{1}{|\mathcal{F}_i|}
    \sum_{e\in\mathcal{F}_i}
    h_{i,e}
    \right].
    \label{eq:recovery_loss}
\end{equation}
This loss preserves what the evidence does for the answer decision, rather than only what the rationale confirms lexically.

\subsection{Overall Objective and Inference}
\label{sec:overall_objective}
The final objective is 
\begin{equation}
    \mathcal{L}_{\mathrm{ours}}
    =
    \mathcal{L}_{\mathrm{OST}}
    +
    \lambda_{\mathrm{rat}}\mathcal{L}_{\mathrm{rat}}
    +
    \lambda_{\mathrm{rec}}\mathcal{L}_{\mathrm{rec}},
    \label{eq:overall_loss}
\end{equation}
where $\lambda_{\mathrm{rat}}$ and $\lambda_{\mathrm{rec}}$ are hyperparameters controlling the contribution of the rationale-token and rationale-conditioned recovery losses.
The base loss preserves generic next-token quantization fidelity on the full calibration pool.
The two additional losses preserve answer-supporting rationale behavior on the faithfulness cache.
After optimization, the optimized transformations are fused into the model weights and W4A4KV4 quantization is applied as in OSTQuant.
The faithfulness cache, teacher model, gold labels, quantized probe, and evidence extractor are used only during calibration-time optimization.
At inference time, the deployed model has the same quantization format, architecture, and runtime cost as the OSTQuant baseline.

\section{Experiments}
\label{sec:experiments}
We evaluate whether explanation-aware transformation optimization preserves both medical QA performance and answer-supporting rationales under W4A4KV4 quantization.

\subsection{Experimental Setup}
\label{sec:experimental_setup}

\paragraph{Datasets}
We evaluate on MedExQA~\citep{kim-etal-2024-medexqa}, MedExpQA~\citep{alonso2024medexpqa}, and ChallengeClinicalQA~\citep{chen-etal-2025-benchmarking}.
For quantization calibration, we construct a mixed pool of 128 non-test medical QA examples from the three datasets. The same calibration pool is used for OSTQuant and our method. Our method additionally builds the faithfulness cache from this pool, and the cache is used only during calibration-time transformation optimization. 
Dataset statistics and split details are in Appendix~\ref{sec:dataset_details}.

\paragraph{Models and quantization}
We evaluate OpenBioLLM-8B~\citep{OpenBioLLMs}, Llama-3-8B-Instruct~\citep{grattafiori2024llama}, BioMistral-7B~\citep{labrak2024biomistral}, and Mistral-7B-Instruct~\citep{jiang2023mistral7b}.
For each model, the full-precision version is used as the reference teacher. The primary quantized systems use OSTQuant~\citep{hu2025ostquant} under W4A4KV4 quantization. 
We set $\lambda_{\mathrm{rat}}=\lambda_{\mathrm{rec}}=\lambda$ to keep the hyperparameter search space tractable and select a single global $\lambda=0.1$ on the MedExpQA validation split. This value is fixed for all datasets, models, and seeds. Appendix~\ref{sec:lambda_sensitivity} reports sensitivity. 

As an additional PTQ-method check, we also adapt the explanation-aware objective to OmniQuant~\citep{shao2024omniquant} on Llama-3-8B-Instruct and Mistral-7B-Instruct across the three evaluation datasets. Full quantization, optimization, and hyperparameter settings are reported in Appendix~\ref{sec:quantization_details}.

\paragraph{Compared systems}
We compare three systems.
\textbf{Full precision} is the original unquantized model.
\textbf{OSTQuant} is the same-calibration PTQ baseline using the original OSTQuant objective.
\textbf{Ours} uses the same OSTQuant pipeline, calibration pool, W4A4KV4, and fused inference path, but adds the rationale-token and rationale-conditioned recovery losses during transformation optimization.

\subsection{Evaluation Protocol}
\label{sec:evaluation_protocol}

\paragraph{Answer accuracy and Full-precision agreement}
Accuracy measures whether the selected option matches the gold answer. However, as previously noted, accuracy alone can be misleading under PTQ. Prior work on compressed LLM evaluation shows that aggregate accuracy can hide instance-level answer flips, where compressed models change predictions relative to the baseline even when benchmark accuracy remains similar or improves~\citep{dutta2024accuracy}. We report both answer accuracy and full-precision agreement. Full-precision agreement is the fraction of examples where a quantized system selects the same option as the full-precision reference.

\paragraph{Rationale-based answer reconstruction}
Answer accuracy and reference agreement do not show whether a generated rationale supports the model's own answer. We therefore evaluate predicted-answer reconstruction, a simulatability-style test~\citep{hase2020leakage,wiegreffe-etal-2021-measuring}. For each system, we generate a concise rationale. We then remove direct answer-label cues and ask the same system to recover an answer from the question, answer options, and generated rationale. Predicted-answer reconstruction, or PredRec, is the fraction of examples where the recovered answer matches the system's original selected answer. We also compute gold-answer reconstruction, or GoldRec, where the target is the gold answer. Because PredRec and $\mathcal{L}_{\mathrm{rec}}$ use related forced-choice scoring, we also perform additional control experiments. Appendix~\ref{app:predrec_controls} reports the full control results.

\paragraph{Full-precision evidence retention}
PredRec does not directly measure whether quantized rationales retain the clinical evidence used by the full-precision model. We therefore add Full-precision Evidence Retention (FER), inspired by the atomic-claim support style of FActScore~\citep{min2023factscore}. Unlike FActScore's external factuality evaluation, FER asks whether the system rationale preserves atomic clinical claims from the full-precision rationale for the same question. To focus on evidence retention rather than answer differences, we compute FER only on examples where full-precision, OSTQuant, and our method select the same answer.

For each example, we decompose the full-precision rationale into atomic clinical claims and judge whether each claim is supported by the system rationale in the context of the question and answer options.
Let $\mathcal{C}_i^{\mathrm{fp}}$ be the extracted full-precision claims, and let $J(c,r,q,\mathcal{O})=1$ if claim $c$ is supported by rationale $r$ under question $q$ and options $\mathcal{O}$. We define
\begin{equation}
    \operatorname{FER}(s)
    =
    \frac{1}{N}
    \sum_{i=1}^{N}
    \frac{1}{|\mathcal{C}_i^{\mathrm{fp}}|}
    \sum_{c\in\mathcal{C}_i^{\mathrm{fp}}}
    J(c,r_i^s,q_i,\mathcal{O}_i).
\end{equation}

FER measures retention of full-precision clinical evidence, not absolute medical factuality, and does not penalize additional unsupported claims.
To assess GPT-judge reliability, we conduct a small human validation on 20 Llama-3/MedExpQA cases (126 claims, 252 candidate--claim judgments per annotator). 
Two annotators agree on 79.37\% of judgments (Cohen's $\kappa = 0.58$), and the automatic judge agrees with the pooled human labels on 80.95\% ($\kappa = 0.63$). The automatic judge is stricter in absolute support rate but shows the same directional gap between the two systems ($+9.52$ for humans versus $+11.90$ for the automatic judge). Full details are in Appendix~\ref{app:human_validation}.

\paragraph{Unsupported-Claim Rate}
FER measures recall of evidence from the full-precision rationale, but it does not penalize additional claims introduced by a system rationale. We therefore report the Unsupported-Claim Rate (UCR) as a complementary metric. For each system rationale $r_i^s$, we extract atomic clinical claims and judge each claim relative to the full-precision rationale. The judge labels each claim as source-supported, valid background, unsupported added, contradicted, or not a claim. Let $\mathcal{C}_i^s$ denote the extracted claims receiving a valid claim judgment, excluding items labeled not a claim, and let $U(c,r_i^{\mathrm{fp}},q_i,\mathcal{O}_i)=1$ if claim $c$ is labeled unsupported added or contradicted relative to the full-precision rationale, and $0$ otherwise. We define 
\begin{equation}
\operatorname{UCR}(s)
=
\frac{1}{N}
\sum_{i=1}^{N}
\frac{1}{|\mathcal{C}_i^s|}
\sum_{c\in\mathcal{C}_i^s}
U(c,r_i^{\mathrm{fp}},q_i,\mathcal{O}_i).
\end{equation}
UCR measures the rate of claims that are unsupported by or contradictory to the full-precision rationale, with lower values indicating fewer unsupported additions. Like FER, UCR is reference-based and does not measure absolute medical factuality.

\paragraph{Pairwise rationale judging}
As a supplementary analysis, we conduct an anonymized LLM-as-a-judge pairwise comparison on MedExpQA. We focus on MedExpQA because it shows the largest answer--rationale degradation in Table~\ref{tab:modelwise_answer_rationale}, making rationale-level differences most measurable. We treat this analysis as supporting evidence rather than a primary metric because prior work suggests that LLM judges may miss subtle quantization-induced changes in self-explanations~\citep{wang2026can}. To focus on rationale quality rather than answer differences, we restrict the comparison to cases where OSTQuant and our method select the same answer. The judge receives the question, answer options, selected answer, and two anonymized rationales, and chooses which rationale is more clinically supported. Full counts and qualitative examples are reported in Appendix~\ref{app:pairwise_details} and Appendix~\ref{sec:appendix_qualitative}.

All evaluation prompts are provided in Appendix~\ref{sec:prompts}. All LLM-based evaluations use GPT-5.4 (as of 2026-05).

\section{Results and Analysis}
\label{sec:results}
All main results use $\lambda=0.1$ for OSTQuant and $0.5$ for OmniQuant. Quantized results are reported as mean$\pm$standard deviation over three random seeds (0,1,2). Answer scoring is deterministic and rationales are generated with fixed decoding settings. Full-precision results use one deterministic run.

\subsection{Answer Preservation and Rationale Reconstruction}
\label{sec:main_results}

\begin{table*}[t]
\centering
\scriptsize
\setlength{\tabcolsep}{2.3pt}
\renewcommand{\arraystretch}{1.05}
\begin{tabular*}{\textwidth}{@{\extracolsep{\fill}}llccccc ccc@{}}
\toprule
& & \multicolumn{2}{c}{Full-precision} 
& \multicolumn{3}{c}{OSTQuant} 
& \multicolumn{3}{c}{Ours} \\
\cmidrule(lr){3-4}
\cmidrule(lr){5-7}
\cmidrule(lr){8-10}
Dataset & Model 
& Acc. & PredRec 
& Acc. & FP Agreement & PredRec 
& Acc. & FP Agreement & PredRec \\
\midrule

\multirow{5}{*}{MedExQA}
& OpenBioLLM 
& 38.62 & 86.81 
& $34.82{\pm}1.23$ & $77.77{\pm}2.35$ & $83.48{\pm}0.16$
& $36.45{\pm}1.05$ & $\mathbf{81.03{\pm}0.71}$ & $\mathbf{86.60{\pm}1.81}$ \\

& Llama-3 
& 41.60 & 94.57
& $37.45{\pm}0.28$ & $84.47{\pm}1.39$ & $92.20{\pm}0.64$
& $38.97{\pm}1.14$ & $\mathbf{87.23{\pm}1.15}$ & $\mathbf{93.62{\pm}0.32}$ \\

& BioMistral 
& 58.72 & 78.94
& $55.57{\pm}0.93$ & $\mathbf{78.48{\pm}0.96}$ & $77.52{\pm}3.82$
& $56.17{\pm}1.15$ & $78.19{\pm}0.21$ & $\mathbf{78.23{\pm}2.83}$ \\

& Mistral 
& 64.47 & 94.68
& $60.57{\pm}1.66$ & $79.50{\pm}0.59$ & $87.52{\pm}3.96$
& $60.14{\pm}1.38$ & $\mathbf{79.57{\pm}1.08}$ & $\mathbf{89.86{\pm}1.72}$ \\

\cmidrule(lr){2-10}
& Avg. 
& 50.85 & 88.75
& 47.10 & 80.06 & 85.18
& 47.93 & \textbf{81.50} & \textbf{87.08} \\

\midrule

\multirow{5}{*}{MedExpQA}
& OpenBioLLM 
& 24.80 & 91.20
& $23.20{\pm}2.12$ & $38.40{\pm}3.49$ & $39.73{\pm}5.21$
& $24.00{\pm}1.39$ & $\mathbf{82.13{\pm}6.47}$ & $\mathbf{86.40{\pm}4.23}$ \\

& Llama-3 
& 24.80 & 100.00
& $34.93{\pm}5.45$ & $33.33{\pm}2.44$ & $49.07{\pm}4.41$
& $26.67{\pm}1.85$ & $\mathbf{83.73{\pm}2.44}$ & $\mathbf{99.20{\pm}0.80}$ \\

& BioMistral 
& 34.40 & 76.80
& $41.87{\pm}1.67$ & $63.20{\pm}1.39$ & $57.87{\pm}1.67$
& $32.27{\pm}1.67$ & $\mathbf{76.27{\pm}2.44}$ & $\mathbf{74.67{\pm}3.23}$ \\

& Mistral 
& 43.20 & 93.60
& $47.20{\pm}2.77$ & $59.73{\pm}3.33$ & $79.47{\pm}2.81$
& $43.20{\pm}0.80$ & $\mathbf{75.73{\pm}4.62}$ & $\mathbf{93.60{\pm}4.23}$ \\

\cmidrule(lr){2-10}
& Avg. 
& 31.80 & 90.40
& 36.80 & 48.66 & 56.54
& 31.54 & \textbf{79.47} & \textbf{88.47} \\

\midrule

\multirow{5}{*}{ChallengeClin.}
& OpenBioLLM 
& 25.00 & 90.91
& $24.24{\pm}2.76$ & $76.62{\pm}0.86$ & $86.47{\pm}3.62$
& $24.13{\pm}1.14$ & $\mathbf{79.76{\pm}0.50}$ & $\mathbf{89.07{\pm}3.47}$ \\

& Llama-3 
& 24.03 & 98.38
& $23.16{\pm}0.19$ & $88.31{\pm}0.97$ & $95.24{\pm}0.50$
& $23.81{\pm}0.75$ & $\mathbf{90.80{\pm}1.46}$ & $\mathbf{96.32{\pm}0.50}$ \\

& BioMistral 
& 29.55 & 66.23
& $29.11{\pm}2.30$ & $71.10{\pm}1.17$ & $\mathbf{69.70{\pm}1.23}$
& $29.00{\pm}0.50$ & $\mathbf{71.86{\pm}2.52}$ & $69.48{\pm}5.96$ \\

& Mistral 
& 33.77 & 97.40
& $31.39{\pm}0.37$ & $70.35{\pm}1.67$ & $93.29{\pm}2.48$
& $30.09{\pm}2.62$ & $\mathbf{71.97{\pm}2.16}$ & $\mathbf{95.13{\pm}0.56}$ \\

\cmidrule(lr){2-10}
& Avg. 
& 28.09 & 88.23
& 26.98 & 76.60 & 86.18
& 26.76 & \textbf{78.60} & \textbf{87.50} \\

\midrule
\multicolumn{2}{@{}l}{Overall avg.}
& 36.91 & 89.13
& 36.96 & 68.44 & 75.96
& 35.41 & \textbf{79.86} & \textbf{87.68} \\

\bottomrule
\end{tabular*}
\caption{
Model-wise answer and rationale-preservation results under W4A4KV4 quantization.
Full-precision agreement is 100 by construction and omitted for compactness.
Bold indicates the better quantized system for reference agreement or PredRec.
}
\label{tab:modelwise_answer_rationale}
\end{table*}

Table~\ref{tab:modelwise_answer_rationale} reports three complementary metrics. Accuracy measures correctness with respect to the gold answer. Full-precision agreement measures whether the quantized model preserves the answer chosen by the full-precision reference. PredRec measures whether the generated rationale supports the system's own selected answer.

\paragraph{Accuracy and Full-precision answer agreement}
On MedExpQA, OSTQuant raises average accuracy from 31.80 to 36.80 but reduces full-precision agreement to 48.66, suggesting that many gains reflect quantization-induced answer shifts rather than preservation of reference behavior. Our method keeps accuracy close to the full-precision reference and improves agreement to 79.47. Overall agreement also rises from 68.44 to 79.86.

\paragraph{Predicted-answer reconstruction}
PredRec reveals the strongest rationale-level effect. On MedExpQA, OSTQuant's average PredRec drops from the full-precision value of 90.40 to 56.54, even though its accuracy increases. \emph{This means that a quantized model can achieve favorable answer accuracy while producing rationales that no longer support its selected answers}. Our method substantially reduces this failure, close to the reference. On MedExQA and ChallengeClinicalQA, OSTQuant causes less severe rationale degradation, but our method remains comparable or slightly better on average. These results support the central claim that preserving or improving final-answer accuracy does not guarantee preservation of answer-supporting rationales. Because our objective is preservation of full-precision behavior rather than maximization of gold-label accuracy, we treat accuracy as a secondary outcome. The higher OSTQuant accuracy on MedExpQA should therefore not be interpreted as stronger preservation.

\subsection{Evidence Retention and Unsupported Claims}
FER and UCR capture complementary rationale drift on MedExpQA same-prediction examples: FER measures retention of full-precision evidence, while UCR measures unsupported or contradictory additions.

Table~\ref{tab:fer_ucr} reports paired differences for FER and UCR on MedExpQA same-prediction examples. For FER, our method improves three of four model-level point estimates and raises the model average by $+5.34$ points, with a 95\% CI of $[1.50, 9.18]$. For UCR, lower is better; our method lowers all four point estimates and reduces the model average by $-4.65$ points, with a 95\% CI of $[-9.10, -0.28]$. 
BioMistral is the only FER exception. This likely reflects its already-low full-precision PredRec (76.8, the lowest among the four models) and weaker rationale structure, leaving less answer-supporting behavior to preserve. Its UCR still decreases, suggesting fewer unsupported additions despite no evidence-retention gain.
These average intervals exclude zero, but most model-specific intervals do not, so we interpret the results as complementary evidence of reduced rationale drift on average rather than uniform per-model improvement. Both metrics remain reference-based and do not establish absolute medical factuality. 

\begin{table}[t]
\centering
\footnotesize
\setlength{\tabcolsep}{2pt}
\resizebox{\columnwidth}{!}{%
\begin{tabular}{@{}lcc@{}}
\toprule
Model & FER $\Delta$ $\uparrow$ & UCR $\Delta$ $\downarrow$ \\
\midrule
OpenBioLLM & $+5.92$ [$-3.09$, 14.94]
& $-7.15$ [$-16.65$, $+2.37$] \\
Llama-3 & $+12.35$ [8.77, 15.94]
& $-2.35$ [$-13.62$, $+8.70$] \\
BioMistral & $-1.07$ [$-2.14$, 0.01]
& $-6.56$ [$-13.96$, $+1.01$] \\
Mistral & $+4.15$ [$-6.32$, 14.63]
& $-2.53$ [$-8.51$, $+3.68$] \\
\midrule
Avg. & $+5.34$ [1.50, 9.18]
& $-4.65$ [$-9.10$, $-0.28$] \\
\bottomrule
\end{tabular}%
}
\caption{FER and UCR differences on MedExpQA same-prediction
examples. Values are Ours minus OSTQuant with 95\% CIs. Positive
favors Ours for FER, negative favors Ours for UCR.}
\label{tab:fer_ucr}
\end{table}

\subsection{Additional Check with OmniQuant}
\label{sec:omniquant_check}
We additionally adapt the objective to OmniQuant for Llama-3-8B-Instruct and Mistral-7B-Instruct. Table~\ref{tab:omniquant} shows answer-behavior and rationale-support drift under OmniQuant. Adding our objective improves full-precision agreement and PredRec on all three datasets and partially recovers accuracy. Although the gains are smaller than in the OSTQuant setting, the results suggest that the proposed objective can improve rationale preservation beyond OSTQuant, with effectiveness depending on the PTQ pipeline and optimization interface.

\begin{table}[t]
\centering
\footnotesize
\renewcommand{\arraystretch}{1.06}
\setlength{\tabcolsep}{3.5pt}

\resizebox{\columnwidth}{!}{%
\begin{tabular}{@{}lccc@{}}
\toprule
System & Acc. & FP Ag. & PredRec \\
\midrule

\multicolumn{4}{@{}l}{\textbf{Llama-3-8B-Instruct}} \\[2pt]

\multicolumn{4}{@{}l}{\textit{MedExQA}} \\[1pt]
\quad OmniQuant
& $23.37{\pm}0.34$
& $28.79{\pm}2.29$
& $47.91{\pm}3.47$ \\
\quad Ours
& $29.50{\pm}0.62$
& $\mathbf{55.67{\pm}1.55}$
& $\mathbf{57.45{\pm}0.98}$ \\

\cmidrule(lr){1-4}

\multicolumn{4}{@{}l}{\textit{MedExpQA}} \\[1pt]
\quad OmniQuant
& $22.13{\pm}3.03$
& $23.73{\pm}2.01$
& $51.47{\pm}2.01$ \\
\quad Ours
& $24.27{\pm}3.33$
& $\mathbf{53.87{\pm}5.45}$
& $\mathbf{58.13{\pm}7.43}$ \\

\cmidrule(lr){1-4}

\multicolumn{4}{@{}l}{\textit{ChallengeClin.}} \\[1pt]
\quad OmniQuant
& $20.78{\pm}2.34$
& $24.03{\pm}2.58$
& $48.59{\pm}1.60$ \\
\quad Ours
& $23.81{\pm}1.63$
& $\mathbf{50.97{\pm}2.03}$
& $\mathbf{57.58{\pm}2.72}$ \\

\midrule

\multicolumn{4}{@{}l}{\textbf{Mistral-7B-Instruct}} \\[2pt]

\multicolumn{4}{@{}l}{\textit{MedExQA}} \\[1pt]
\quad OmniQuant
& $30.28{\pm}2.72$
& $52.76{\pm}3.54$
& $53.37{\pm}2.43$ \\
\quad Ours
& $37.91{\pm}1.29$
& $\mathbf{62.52{\pm}0.82}$
& $\mathbf{65.82{\pm}0.69}$ \\

\cmidrule(lr){1-4}

\multicolumn{4}{@{}l}{\textit{MedExpQA}} \\[1pt]
\quad OmniQuant
& $25.60{\pm}2.88$
& $47.91{\pm}4.55$
& $49.87{\pm}4.03$ \\
\quad Ours
& $31.20{\pm}2.40$
& $\mathbf{66.36{\pm}3.48}$
& $\mathbf{66.93{\pm}0.92}$ \\

\cmidrule(lr){1-4}

\multicolumn{4}{@{}l}{\textit{ChallengeClin.}} \\[1pt]
\quad OmniQuant
& $21.97{\pm}3.69$
& $44.96{\pm}4.17$
& $45.24{\pm}1.60$ \\
\quad Ours
& $27.06{\pm}2.93$
& $\mathbf{57.31{\pm}4.41}$
& $\mathbf{58.23{\pm}4.62}$ \\

\bottomrule
\end{tabular}%
}

\caption{
Two-model OmniQuant check. Bold marks the better quantized system for FP agreement and PredRec.
}
\label{tab:omniquant}
\end{table}

\section{Ablation Study}
\label{sec:ablation}
We conduct ablations on Llama-3-8B-Instruct with MedExpQA. Accuracy and PredRec are mean$\pm$standard deviation over three seeds, while FER is computed for one fixed seed on the corresponding same-prediction subset and reported with a 95\% confidence interval. For FER, the same-prediction subset is constructed separately for each ablation table, so FER values are directly comparable within a table but may differ for the same setting across tables.

\paragraph{Simple Rationale Baselines}
Table~\ref{tab:simple_baselines} compares against full-rationale KD and random span selection. All rationale-aware variants improve PredRec over OSTQuant, but they also reduce task accuracy, showing the accuracy--preservation trade-off in this setting. Among the rationale-aware variants, our method has comparable accuracy, the highest PredRec, and the highest FER. This suggests that evidence selection provides benefit beyond generic rationale distillation or arbitrary span selection.

\begin{table}[t]
\centering
\footnotesize
\setlength{\tabcolsep}{0pt}
\renewcommand{\arraystretch}{1.02}
\begin{tabular*}{\linewidth}{@{\extracolsep{\fill}}lccc@{}}
\toprule
System & Acc. & PredRec & FER \\
\midrule
OSTQuant
& $34.93{\pm}5.45$
& $49.07{\pm}4.41$
& \cival{53.25}{[45.90, 60.58]} \\
Full KD
& $27.20{\pm}2.12$
& $95.20{\pm}1.60$
& \cival{56.30}{[50.95, 61.51]} \\
Random
& $24.80{\pm}0.80$
& $97.87{\pm}1.22$
& \cival{55.99}{[51.68, 60.12]} \\
Ours
& $26.67{\pm}1.85$
& $\mathbf{99.20{\pm}0.80}$
& \cival{\textbf{59.15}}{\textbf{[54.68, 63.51]}} \\
\bottomrule
\end{tabular*}
\caption{Simple rationale baselines on Llama-3/MedExpQA.}
\label{tab:simple_baselines}
\end{table}

\paragraph{Loss Component Ablation}
\label{sec:loss_component_ablation}
Table~\ref{tab:loss_ablation} ablates the two auxiliary losses. Either loss improves PredRec over the base objective, while accuracy remains in a similar range across the auxiliary-loss variants. FER separates their effects: $\mathcal{L}_{\rm rec}$ alone retains more full-precision evidence than $\mathcal{L}_{\rm rat}$ alone, and the combined objective obtains the highest observed FER and PredRec. This suggests that token-level preservation and rationale-conditioned answer recovery are complementary.
\begin{table}[t]
\centering
\footnotesize
\setlength{\tabcolsep}{0pt}
\renewcommand{\arraystretch}{1.02}
\begin{tabular*}{\linewidth}{@{\extracolsep{\fill}}lccc@{}}
\toprule
Objective & Acc. & PredRec & FER \\
\midrule
Base
& $34.93{\pm}5.45$
& $49.07{\pm}4.41$
& \cival{51.64}{[40.41, 62.00]} \\
$+\mathcal{L}_{\mathrm{rat}}$
& $24.80{\pm}0.80$
& $95.73{\pm}2.44$
& \cival{46.01}{[36.36, 55.79]} \\
$+\mathcal{L}_{\mathrm{rec}}$
& $24.27{\pm}1.22$
& $96.27{\pm}2.01$
& \cival{53.99}{[45.93, 62.37]} \\
Both
& $26.67{\pm}1.85$
& $\mathbf{99.20{\pm}0.80}$
& \cival{\textbf{58.22}}{\textbf{[49.33, 66.82]}} \\
\bottomrule
\end{tabular*}
\caption{Loss-component ablation on Llama-3/MedExpQA.}
\label{tab:loss_ablation}
\end{table}

\paragraph{Evidence Selection Ablation}
\label{sec:evidence_selection_ablation}
Table~\ref{tab:selection_ablation} compares cumulative cache variants. PredRec is already high with all proposed spans, so additional filtering is not well distinguished by PredRec alone. FER shows that the largest gain appears after adding the quantization-weakened filter, and the full margin-drop criterion gives the highest observed FER and PredRec. Thus, the probe-based filtering stages help focus the cache on evidence most affected by quantization while maintaining strong rationale-to-answer consistency.

\begin{table}[t]
\centering
\footnotesize
\setlength{\tabcolsep}{0pt}
\renewcommand{\arraystretch}{1.02}
\begin{tabular*}{\linewidth}{@{\extracolsep{\fill}}lccc@{}}
\toprule
Cache & Acc. & PredRec & FER \\
\midrule
All spans
& $25.33{\pm}2.01$
& $97.60{\pm}1.39$
& \cival{51.31}{[46.58, 56.16]} \\
$+$ support
& $24.80{\pm}1.60$
& $97.07{\pm}1.67$
& \cival{51.15}{[45.53, 56.75]} \\
$+$ weakened
& $25.33{\pm}2.31$
& $95.47{\pm}3.03$
& \cival{54.99}{[49.69, 60.09]} \\
$+$ margin
& $26.67{\pm}1.85$
& $\mathbf{99.20{\pm}0.80}$
& \cival{\textbf{55.30}}{\textbf{[50.41, 60.31]}} \\
\bottomrule
\end{tabular*}
\caption{Evidence-selection ablation on Llama-3/MedExpQA.}
\label{tab:selection_ablation}
\end{table}

\section{Conclusion}
We studied whether low-bit PTQ preserves the evidence behind medical QA answers. Our results show that a calibration-matched OSTQuant baseline can keep aggregate accuracy similar while weakening full-precision agreement and rationale-to-answer support, suggesting that accuracy alone may mask explanation-level degradation. Our explanation-aware objective uses full-precision teacher rationales to preserve evidence-token and evidence-conditioned answer behavior during calibration. Across four 7B--8B models and three benchmarks, it better preserves answer behavior and answer-supporting rationales. The OmniQuant check further suggests broader use as a preservation regularizer, although recovery remains method-dependent. Overall, these findings motivate evaluating PTQ with evidence-preservation metrics, not only final-answer accuracy.

\section*{Limitations}
This study focuses on medical multiple-choice QA and does not evaluate open-ended clinical generation, real-world clinical decision support, or patient-facing deployment. Although our method improves answer-rationale consistency, it does not guarantee clinical correctness, factuality, or complete explanation faithfulness. The faithfulness cache is built from full-precision teacher rationales, so it can inherit teacher errors or preserve teacher-specific reasoning patterns. The prompted extractor is used only as a candidate span proposer and is followed by behavior-based filtering, but extraction errors can still affect which evidence receives preservation pressure.

Our full evaluation primarily instantiates the method on OSTQuant under W4A4KV4 quantization with four 7B--8B models. We include an OmniQuant generalization check, but broader conclusions may differ for other PTQ backbones, bit-widths, model sizes, calibration pools, or non-medical domains. The pairwise LLM-as-a-judge analysis may also reflect judge-model preferences despite anonymization, a fixed rubric, and an explicit tie option. Finally, this work should be viewed as a compression and evaluation study for explanation-critical settings, not as evidence that any quantized model is safe for clinical use.

\section*{Ethical Considerations}
This work studies compression and explanation preservation for medical QA models, but it is not intended for clinical decision-making or patient-facing deployment. Quantized models may preserve final answers while altering the rationale evidence supporting those answers, which could affect user trust and interpretation in high-stakes settings. Our results therefore highlight the importance of evaluating evidence preservation and rationale consistency; however, such evaluation does not guarantee clinical safety, factual correctness, or deployment readiness. All experiments are conducted offline on benchmark datasets intended for research use. We used AI assistants for limited writing and editorial support, such as improving clarity and readability. All technical content, experiments, analyses, and reported results were produced and verified by the authors.

\section*{Acknowledgements}
This work was supported by the Natural Sciences and Engineering Research Council of Canada (NSERC) Collaborative Research and Training Experience (CREATE) program “From Data to Decision” (FD2D). This research was  supported by the Alberta Machine Intelligence Institute (Amii), NSERC (including grants DGECR-2022-00369, RGPIN-2022-03469, and RGPIN-2025-05572), and Alberta Innovates (Enabling Better Health through Artificial Intelligence (AI-Better Health) Program). We thank Brett Siemens and Miron Nekhoroshkov for assisting with the human validation study.

\bibliography{custom}

\appendix

\section{Artifact licenses and terms}
We use all artifacts only for non-redistributive offline research experiments. MedExQA is released under CC BY-NC-SA 4.0, and MedExpQA is released under CC BY 4.0. For ChallengeClinicalQA, we use the released Medbullets files from the public repository. The evaluated model checkpoints are used under their corresponding terms: OpenBioLLM-8B and Llama-3-8B-Instruct under the Meta Llama 3 license, BioMistral-7B and Mistral-7B-Instruct-v0.1, and Gemma-4-12B-IT under Apache-2.0, and Qwen2.5-72B-Instruct under the Qwen license. OSTQuant and OmniQuant are used under Apache-2.0 and MIT license, respectively, and GPT-based evaluation is performed through the OpenAI API under the applicable service terms.

\section{Dataset and Calibration Details}
\label{sec:dataset_details}
Table~\ref{tab:dataset_statistics_appendix} summarizes datasets, evaluation splits, and calibration-pool composition. All calibration examples are sampled from non-test portions of the corresponding datasets. No test example is used for transformation optimization, cache construction, hyperparameter selection, or prompt development. For ChallengeClinicalQA, we use only the Medbullets subset, and the calibration pool is drawn from the four-option Medbullets subset and evaluation is performed on the disjoint five-option Medbullets subset.

\begin{table}[t]
\centering
\small
\setlength{\tabcolsep}{3.5pt}
\renewcommand{\arraystretch}{1.05}

\resizebox{\columnwidth}{!}{
\begin{tabular}{lcccc}
\toprule
Dataset & Train/Pool & Dev & Test & Calib. used \\
\midrule
MedExQA & -- & 25 & 940 & 25 \\
MedExpQA & 434 & 63 & 125 & 51 \\
ChallengeClin. & 308 (4-opt.) & -- & 308 (5-opt.) & 52 \\
\bottomrule
\end{tabular}
}

\caption{
Dataset statistics and calibration-pool composition.
A total of 128 non-test calibration examples were used.
}
\label{tab:dataset_statistics_appendix}
\end{table}

\paragraph{Calibration-pool construction}
We sample 128 non-test examples from the three datasets. The same 128 examples are used by OSTQuant and our method. OSTQuant uses this pool for the base transformation-optimization loss. Our method uses the same pool for the base loss and additionally builds a faithfulness cache.

\section{Quantization and Optimization Details}
\label{sec:quantization_details}
This section reports the quantization and calibration-time optimization settings for the primary OSTQuant experiments and the additional OmniQuant check. OSTQuant and our method use the same calibration pool, deployed W4A4KV4 format, and fused inference path. The OmniQuant check uses the same mixed medical calibration pool, but evaluates a W4A4 fake-quantized inference path, where the weights and activations are quantized-dequantized during forward passes using learned OmniQuant parameters.

\paragraph{Shared explanation-aware objective settings}
Table~\ref{tab:shared_objective_settings} lists the settings for the faithfulness cache and the explanation-aware objective. These components are used only during cache construction and calibration-time optimization, and are discarded before inference. The only objective hyperparameters that differ between the OSTQuant-backed and OmniQuant-backed runs are the two auxiliary loss weights.

\begin{table}[t]
\centering
\small
\setlength{\tabcolsep}{4pt}
\renewcommand{\arraystretch}{1.05}
\begin{tabularx}{\columnwidth}{@{}p{0.65\columnwidth}X@{}}
\toprule
Item & Value \\
\midrule
Rationale-token loss weight $\lambda_{\mathrm{rat}}$
& 0.1 for OSTQuant, 0.5 for OmniQuant \\

Recovery loss weight $\lambda_{\mathrm{rec}}$
& 0.1 for OSTQuant, 0.5 for OmniQuant \\

Faithfulness-cache max sequence length
& 640 \\

Teacher-answer-support threshold $\eta_F$
& 0.02 \\

Probe-weakening threshold $\eta_Q$
& 0.02 \\

Evidence-support sufficiency weight $\alpha$
& 1.0 \\

Margin-drop weight $\beta$
& 0.01 \\

Span proposer
& Qwen2.5-72B-Instruct \\

Maximum evidence units
& 6 \\

Maximum evidence words
& 20 \\

Minimum evidence words
& 3 \\

\bottomrule
\end{tabularx}
\caption{
Shared settings for the explanation-aware objective and faithfulness cache.
Only the auxiliary loss weights differ between the OSTQuant-backed and OmniQuant-backed runs.
}
\label{tab:shared_objective_settings}
\end{table}

\paragraph{OSTQuant-backed primary experiments}
Table~\ref{tab:ostquant_settings_appendix} lists the quantization and calibration-time optimization settings shared by OSTQuant and our OSTQuant-backed method. The sequence length and calibration-sample count refer to calibration-time transformation optimization, not to a restriction on downstream inference length.

\begin{table}[t]
\centering
\small
\setlength{\tabcolsep}{4pt}
\renewcommand{\arraystretch}{1.05}
\begin{tabularx}{\columnwidth}{@{}p{0.43\columnwidth}X@{}}
\toprule
Item & Value \\
\midrule
Evaluated format
& W4A4KV4 \\

Calibration data
& 128 mixed non-test medical QA examples, sequence length 512 \\

Base objective
& Top-$K$ KL matching between the quantization-disabled transformed path and the fake-quantized path \\

Implementation loss
& \texttt{kl\_top}, \texttt{post\_attn=True} \\

Weight quantization
& GPTQ, 4-bit symmetric, group size $-1$, MSE-based clipping \\

GPTQ damping
& 0.01 \\

Activation quantization
& 4-bit dynamic per-token asymmetric \\

K/V cache quantization
& 4-bit, group size 128 \\

Unquantized paths
& Residual, attention, and output paths kept in 16-bit, following OSTQuant \\

Computation dtype
& BF16 \\

Optimized transformations
& Global rotation, online Q/K Hadamard, O/V rotation, and smoothing \\

Smoothing modules
& Q/K, O/V, up/down, and norm-linear smoothing \\

Optimizer
& Riemannian Adam \\

Optimization steps
& Maximum 100 \\

Learning rates
& 0.0169 for rotation, 0.00179 for smoothing \\

Smoothing momentum
& 0.9 \\

Gradient checkpointing
& Enabled \\
\bottomrule
\end{tabularx}
\caption{
Quantization and calibration-time optimization settings for the OSTQuant primary experiments.
Both OSTQuant and our method use the same W4A4KV4 format, calibration pool, and fused inference path.
}
\label{tab:ostquant_settings_appendix}
\end{table}

\paragraph{OmniQuant-backed additional check}
Table~\ref{tab:omniquant_settings_appendix} lists the settings for the OmniQuant baseline and our OmniQuant-backed variant. This check is intended as a PTQ-method transfer check for the preservation objective, not as a runtime-equivalent comparison to the packed W4A4KV4 OSTQuant setting.

\begin{table}[t]
\centering
\small
\setlength{\tabcolsep}{4pt}
\renewcommand{\arraystretch}{1.05}
\begin{tabularx}{\columnwidth}{@{}p{0.43\columnwidth}X@{}}
\toprule
Item & Value \\
\midrule
Quantization format
& W4A4  \\

Calibration data
& 128 mixed non-test medical QA examples, sequence length 512 \\

Base objective
& Layer-wise block-output reconstruction \\

Implementation loss
& MSE between full-precision and fake-quantized layer outputs \\

Weight quantization
& 4-bit asymmetric per-channel affine, no group size \\

Learned weight clipping
& Enabled (LWC) \\

Activation quantization
& 4-bit dynamic per-token asymmetric \\

Attention operand quantization
& Q/K/V matmul operands dynamically fake-quantized to 4-bit \\

K/V cache quantization
& No separate packed K/V-cache quantization; cache stored in the FP16 path \\

Compute and attention
& FP16 / AMP, \texttt{eager} attention, attention probabilities kept 16-bit \\

Optimized transformations
& LET smoothing scales and LWC clipping \\

Smoothing modules
& QKV, V/O, Q/K, and MLP up/gate smoothing; shifts disabled for LLaMA-style models \\

Optimizer
& AdamW \\

Optimization schedule
& 20 epochs, batch size 1, 128 steps/epoch, 2560 updates per layer \\

Learning rates
& 0.005 for LET, 0.01 for LWC \\

Weight decay
& 0 \\

LET initialization $\alpha$
& 0.5 \\

Activation statistics
& Precomputed per model/seed and reused \\

Auxiliary-loss integration
& Logit-KL auxiliary loss every 32 OmniQuant optimization steps, auxiliary batch size 4 \\
\bottomrule
\end{tabularx}
\caption{
Quantization and calibration-time optimization settings for the OmniQuant check.
The baseline uses only the layer-wise reconstruction objective, while our variant adds the explanation-aware auxiliary losses using the shared settings in Table~\ref{tab:shared_objective_settings}.
}
\label{tab:omniquant_settings_appendix}
\end{table}

\section{Computational Budget and Infrastructure}
\label{sec:compute_budget}
All local experiments were run on NVIDIA L40S GPUs with 48GB memory. 
Quantization and transformation-optimization runs generally used 4 GPUs per run, while rationale generation, answer scoring, and label-recovery jobs used 1 GPU per run. Qwen2.5-72B-Instruct extraction stages were run locally with HuggingFace inference, typically on 4 GPUs. LLM-as-a-judge evaluation stages, including FER, UCR, and pairwise rationale judging, used the GPT-5.4 API.

\paragraph{Representative calibration cost}
To make the additional calibration cost of our method explicit, Table~\ref{tab:calibration_cost} reports one representative Llama-3-8B-Instruct run. Standard OSTQuant requires 3.11 GPU-hours. Our two-pass procedure first uses the same-calibration OSTQuant run to obtain the fixed quantized probe $M_p$, then spends 1.66 GPU-hours on span proposal and cache construction, followed by 7.83 GPU-hours for the cache-conditioned transformation optimization. The complete procedure therefore requires 12.60 GPU-hours, corresponding to an additional 9.49 GPU-hours relative to standard OSTQuant.

\begin{table}[t]
\centering
\small
\setlength{\tabcolsep}{5pt}
\begin{tabular}{@{}lr@{}}
\toprule
Stage & GPU-hours \\
\midrule
Standard OSTQuant / probe construction & 3.11 \\
Span proposal and cache construction & 1.66 \\
Cache-conditioned optimization & 7.83 \\
\midrule
Ours total & 12.60 \\
Additional cost over OSTQuant & +9.49 \\
\bottomrule
\end{tabular}
\caption{
Representative calibration cost for one
Llama-3-8B-Instruct run. GPU-hours are computed as the
number of allocated GPUs multiplied by wall-clock runtime.
All additional computation is incurred only during
calibration.
}
\label{tab:calibration_cost}
\end{table}

\paragraph{Reuse and inference cost}
The quantized probe and faithfulness cache are constructed once per model and reused across random seeds. Subsequent seeds therefore require only the final cache-conditioned optimization stage. LLM-as-a-judge evaluations are performed through the GPT-5.4 API and are not included in the local GPU-hour totals. All probe, teacher, extractor, and cache components are discarded after calibration. Consequently, the deployed model retains the same W4A4KV4 format, architecture, and inference cost as the OSTQuant baseline.

\paragraph{Study-wide budget}
Based on Slurm accounting, the completed experiments reported in this paper used approximately 3.0k L40S GPU-hours, computed as allocated GPUs multiplied by wall-clock runtime. This total includes the main OSTQuant experiments, ablations, additional OmniQuant checks, rationale generation, cache construction, and supporting analyses run locally.

\section{Lambda Sensitivity}
\label{sec:lambda_sensitivity}
We tune the shared explanation-aware loss weight $\lambda$, where $\lambda_{\mathrm{rat}}=\lambda_{\mathrm{rec}}=\lambda$, on the MedExpQA validation split.
The validation split is small, so we use this analysis only to choose a single global hyperparameter, not as an independent result. As shown in Table~\ref{tab:lambda_sensitivity}, we select $\lambda=0.1$ since it gives the best average validation PredRec while keeping accuracy in the same range as the other settings. The selected value is fixed for all main experiments.

\begin{table}[t]
\centering
\small
\setlength{\tabcolsep}{5pt}
\renewcommand{\arraystretch}{1.03}
\begin{tabular}{ccc}
\toprule
$\lambda$ & Avg. Acc. & Avg. PredRec \\
\midrule
0.01 & 34.39 & 88.63 \\
0.05 & 35.32 & 88.21 \\
0.10 & 34.52 & \textbf{90.21} \\
0.30 & 35.10 & 89.29 \\
0.50 & 35.85 & 88.36 \\
\bottomrule
\end{tabular}
\caption{
Sensitivity to the shared explanation-aware loss weight on the MedExpQA validation split.
Values are averaged over the four evaluated models. Main experiments use $\lambda=0.1$.
}
\label{tab:lambda_sensitivity}
\end{table}

\section{PredRec Content Controls}
\label{app:predrec_controls}
Because PredRec and $\mathcal{L}_{\mathrm{rec}}$ use a related forced-choice option-scoring form, we evaluate whether PredRec depends on rationale content or mainly reflects the recovery protocol. We construct several controls on MedExpQA. Standard recovery uses the question, answer options, and system rationale. Question-only recovery removes the rationale. Rationale-only recovery removes the question and uses the answer options and system rationale. Shuffled recovery replaces the rationale with one from another example. External-reconstructor recovery uses the same standard input setting, but replaces the original recovery model with a fixed Gemma-4-12B-IT~\citep{gemmateam2026gemma4} reconstructor that is not one of the evaluated quantized systems and is not optimized by our method.

\begin{table}[t]
\centering
\footnotesize
\setlength{\tabcolsep}{2.6pt}
\renewcommand{\arraystretch}{1.03}
\begin{tabular}{@{}lrrrrr@{}}
\toprule
System & Std. & Q-only & R-only & Shuf. & Ext. \\
\midrule
OSTQuant & 56.54 & 39.60 & 50.07 & 35.62 & 30.20 \\
Ours     & 88.47 & 75.73 & 83.87 & 71.22 & 64.13 \\
\bottomrule
\end{tabular}
\caption{PredRec content and recovery controls on MedExpQA,
averaged over the four evaluated models. Std. denotes the
standard PredRec setting. Ext. uses Gemma-4-12B-IT as a fixed
external reconstructor under the standard input setting.}
\label{tab:predrec_controls}
\end{table}

The controls show that PredRec is affected by the available rationale signal rather than reflecting only the forced-choice protocol. Removing the rationale or replacing it with a shuffled rationale lowers reconstruction scores for both systems, while the non-trivial question-only and rationale-only scores show that both the question/options and rationale carry recovery signal. The external-reconstructor control gives the same directional comparison, with ours higher than OSTQuant under Gemma-4-12B-IT recovery. We therefore use PredRec as a protocol-aligned consistency metric and interpret it together with FER, UCR, and pairwise judgments.

\section{Gold-Answer Reconstruction}
\label{sec:appendix_gold_reconstruction}
Gold-answer reconstruction asks whether a rationale supports the correct answer. We treat GoldRec as complementary because it is coupled to task correctness, whereas PredRec isolates answer-rationale consistency.

\begin{table}[t]
\centering
\scriptsize
\setlength{\tabcolsep}{2.8pt}
\renewcommand{\arraystretch}{1.02}

\resizebox{\columnwidth}{!}{
\begin{tabular}{llccc}
\toprule
Dataset & Model & FP & OSTQuant & Ours \\
\midrule

MedExQA & OpenBio & 39.79 & $33.48{\pm}1.93$ & $\mathbf{35.57{\pm}1.45}$ \\
MedExQA & Llama & 45.43 & $38.94{\pm}0.49$ & $\mathbf{40.21{\pm}0.95}$ \\
MedExQA & BioMistral & 50.96 & $50.18{\pm}1.40$ & $\mathbf{50.78{\pm}3.57}$ \\
MedExQA & Mistral & 62.98 & $56.67{\pm}2.03$ & $\mathbf{56.91{\pm}2.40}$ \\

\midrule

MedExpQA & OpenBio & 25.60 & $23.20{\pm}0.80$ & $\mathbf{24.27{\pm}0.92}$ \\
MedExpQA & Llama & 24.80 & $\mathbf{27.20{\pm}2.40}$ & $26.40{\pm}1.39$ \\
MedExpQA & BioMistral & 28.80 & $\mathbf{31.47{\pm}4.41}$ & $30.13{\pm}3.95$ \\
MedExpQA & Mistral & 44.00 & $40.00{\pm}2.12$ & $\mathbf{43.47{\pm}1.85}$ \\

\midrule

ChallengeClin.QA & OpenBio & 25.32 & $23.59{\pm}2.09$ & $\mathbf{23.16{\pm}0.82}$ \\
ChallengeClin.QA & Llama & 25.00 & $23.16{\pm}0.68$ & $\mathbf{23.59{\pm}1.63}$ \\
ChallengeClin.QA & BioMistral & 25.00 & $25.54{\pm}1.60$ & $\mathbf{26.62{\pm}0.56}$ \\
ChallengeClin.QA & Mistral & 33.77 & $\mathbf{30.74{\pm}1.79}$ & $29.76{\pm}2.91$ \\
\bottomrule
\end{tabular}
}

\caption{
Gold-answer reconstruction under W4A4KV4 quantization.
Ours uses the $\lambda=0.1$ setting.
}
\label{tab:gold_reconstruction_appendix}
\end{table}

\section{Diagnostic Analysis Details}
\label{sec:diagnostic_details}
This section describes how Table~\ref{tab:loss_cue_alignment} is computed. 
This diagnostic analysis is used only to motivate the method.

\paragraph{Setting}
We run the diagnostic on Llama-3-8B-Instruct with MedExpQA. Let $\mathcal{I}_{\mathrm{diag}}$ be the set of examples where the full-precision model $M_0$ and the OSTQuant model $M_{\mathrm{OST}}$ select the same answer. We use full-precision answer-conditioned rationales as fixed teacher-forced sequences, so the analysis measures rationale-token distribution drift rather than answer-switch cases.

\paragraph{Token-level signal}
For each example $i\in\mathcal{I}_{\mathrm{diag}}$, let $r_i=(r_{i,1},\ldots,r_{i,T_i})$ be the full-precision rationale and let $c_i$ be the rationale-generation prompt. At each rationale-token position $t$, we compute 
\begin{equation}
    d_{i,t}
    =
    \KL\!\left(
    p_{M_0}(\cdot\mid c_i,r_{i,<t})
    \,\Vert\,
    p_{M_{\mathrm{OST}}}(\cdot\mid c_i,r_{i,<t})
    \right).
\end{equation}
This teacher-forced KL measures how much quantization changes the next-token distribution at that rationale position.

\paragraph{Token grouping}
Each rationale token is assigned to one group using the following priority: numeric cues, answer/option tokens, clinical cues, and generic/discourse tokens. Numeric cues include numbers, lab values, dosages, ages, measurements, and other numeric quantities and selected-option text. Clinical cues are question-grounded context phrases that also appear in the full-precision rationale after normalization, excluding stopwords, generic words, numeric spans, and option strings. All remaining tokens are labeled generic/discourse.

\paragraph{Aggregation}
Let $z_{i,t}$ be the group assigned to token $r_{i,t}$, and let $\mathbf{1}[\cdot]$ denote the indicator function. For token group $g$, we report 
\begin{align}
    T(g)
    &=
    \frac{
    \sum_{i\in\mathcal{I}_{\mathrm{diag}}}\sum_{t=1}^{T_i}
    \mathbf{1}[z_{i,t}=g]
    }{
    \sum_{i\in\mathcal{I}_{\mathrm{diag}}}T_i
    }, \\
    S(g)
    &=
    \frac{
    \sum_{i\in\mathcal{I}_{\mathrm{diag}}}\sum_{t=1}^{T_i}
    \mathbf{1}[z_{i,t}=g]d_{i,t}
    }{
    \sum_{i\in\mathcal{I}_{\mathrm{diag}}}\sum_{t=1}^{T_i}d_{i,t}
    }.
\end{align}
We report $T(g)$ and $S(g)$ as TokenPct and SignalPct percentages in Table~\ref{tab:loss_cue_alignment}.
TokenPct measures how often each group appears in full-precision rationales, while SignalPct measures how much of the total full-precision-to-OSTQuant KL mass is assigned to that group.

\section{Faithfulness Cache Details}
\label{sec:cache_algorithm_appendix}
This section provides the cache-construction details omitted from Section~\ref{sec:faithfulness_cache}.
The cache is built offline from the shared non-test calibration pool and is used only during calibration-time transformation optimization.
For each calibration example $x_i=(q_i,\mathcal{O}_i,y_i)$, the full-precision teacher $M_0$ first selects an answer $\hat{y}_i^0$ using Eq.~\ref{eq:option_score} and generates an answer-conditioned rationale $r_i=(r_{i,1},\ldots,r_{i,T_i})$. We keep only examples where $\hat{y}_i^0=y_i$. This is a reliability filter. A prompted Qwen2.5-72B-Instruct model then proposes candidate evidence spans $\mathcal{C}_i$ from $r_i$. The proposer is used only for high-recall candidate generation. Evidence support is determined by the behavioral scoring below. Direct answer-label cues are removed before evidence extraction.

\paragraph{teacher-answer-supporting evidence}
Let $a_i=\hat{y}_i^0$ be the teacher target answer. Since retained examples satisfy $\hat{y}_i^0=y_i$, this target equals the gold option on cached examples. Let $u_i(z)$ denote a recovery prompt containing the question, answer options, and rationale content $z$. Let $r_i^{-e}$ denote the rationale with span $e$ removed, and let $\emptyset$ denote a prompt with no rationale evidence. For any model $M$, define
\begin{equation}
    m_i^M(z)=m_{M,i}(u_i(z);a_i),
\end{equation}
where $m_{M,i}$ is the answer margin in Eq.~\ref{eq:answer_margin}. For each candidate
span $e\in\mathcal{C}_i$, we compute necessity- and sufficiency-style support scores:
\begin{equation}
\begin{aligned}
    N_M(e) &= [m_i^M(r_i)-m_i^M(r_i^{-e})]_+,\\
    S_M(e) &= [m_i^M(e)-m_i^M(\emptyset)]_+,
\end{aligned}
\label{eq:cache_ns_scores}
\end{equation}
where $[x]_+=\max(x,0)$. We aggregate the two views with a maximum, 
\begin{equation}
    I_M(e)=\max\{N_M(e),\alpha S_M(e)\},
    \label{eq:cache_importance}
\end{equation}
so a span can be retained if it strongly supports the teacher answer.
The teacher-answer-supporting evidence set is
\begin{equation}
    \mathcal{F}_i=\{e\in\mathcal{C}_i:I_{M_0}(e)>\eta_F\}.
    \label{eq:cache_teacher_faithful}
\end{equation}
We discard examples with empty $\mathcal{F}_i$ or whose concatenated evidence does not recover $a_i$ under $M_0$.

\paragraph{Quantization-affected evidence}
Let $M_p$ be the fixed same-calibration quantized probe. We apply the same support scoring to $M_p$ and keep teacher-answer-supporting spans whose answer-supporting effect is weakened after quantization. For each $e\in\mathcal{F}_i$, define 
\begin{equation}
\begin{aligned}
    D_i^{\mathrm{imp}}(e)
    &= [I_{M_0}(e)-I_{M_p}(e)]_+,\\
    D_i^{\mathrm{mar}}(e)
    &= [m_i^{M_0}(e)-m_i^{M_p}(e)]_+,\\
    D_i(e)
    &= D_i^{\mathrm{imp}}(e)+\beta D_i^{\mathrm{mar}}(e).
\end{aligned}
\label{eq:cache_quantization_effect}
\end{equation}
The quantization-affected evidence set is
\begin{equation}
    \mathcal{Q}_i=\{e\in\mathcal{F}_i:D_i(e)>\eta_Q\}.
    \label{eq:cache_quantization_affected}
\end{equation}
Cache entries with $\mathcal{Q}_i=\emptyset$ are discarded.

\paragraph{Token weights}
We convert span-level scores into token-level weights over the teacher rationale. Let $\operatorname{cov}_i(t)$ be the set of spans in $\mathcal{Q}_i$ that cover token $r_{i,t}$.
We set
\begin{equation}
    \tilde{\omega}_{i,t}
    =
    \sum_{e\in \operatorname{cov}_i(t)} D_i(e),
    \quad
    \omega_{i,t}
    =
    \frac{\tilde{\omega}_{i,t}}
    {\sum_{s=1}^{T_i}\tilde{\omega}_{i,s}}.
    \label{eq:cache_token_weights}
\end{equation}
Tokens not covered by $\mathcal{Q}_i$ receive zero weight. Each retained cache entry is
\[
\xi_i=(x_i,r_i,\mathcal{F}_i,\mathcal{Q}_i,\omega_i)\in\mathcal{A}.
\]
We use $\mathcal{Q}_i$ and $\omega_i$ for the rationale-token loss, and $\mathcal{F}_i$ for the rationale-conditioned recovery loss. The cache parameters $\alpha,\beta,\eta_F,\eta_Q$ are fixed before evaluation and reported in Appendix~\ref{sec:quantization_details}.

\section{Prompts}
\label{sec:prompts}
This appendix lists the prompts used for answer selection, rationale generation, recovery evaluation, faithfulness-cache construction, full-precision Evidence Retention (FER), Unsupported-Claim Rate (UCR), pairwise judging, and qualitative analysis. Unless otherwise stated, generation calls use deterministic decoding. 

\subsection{Common System Prompts}
\label{sec:prompt_system}

\begin{promptbox}{Answer and rationale system prompt}
You are a careful medical exam assistant. Always follow the requested response format exactly.
\end{promptbox}

\begin{promptbox}{Recovery system prompt}
You are a careful evaluator of medical multiple-choice question explanations.

Your task is to infer which option is best supported by the provided explanation.

Important rules:
Base your decision only on the question, options, and explanation.
Do not use outside medical knowledge beyond what is stated or directly implied by the explanation.
If the explanation is weak or incomplete, still choose the option best supported by the explanation.
\end{promptbox}

\begin{promptbox}{Judge system prompt}
You are an expert medical evaluator for clinical multiple-choice QA. Follow the rubric exactly. Output only the requested JSON object.
\end{promptbox}

\subsection{Answer Selection Prompt}
\label{sec:prompt_answer_selection}
For answer selection, we construct the prompt below and score each option completion after \texttt{Final:} using the length-normalized conditional log-likelihood in Eq.~\ref{eq:option_score}. The model is not evaluated by free-form answer parsing.

\begin{promptbox}{Answer selection prompt}
Answer the medical multiple-choice question.
Select the single best answer.

Return the option label followed by the option text.

Question:
{question}

{optional_contexts}

Options:
{label}. {option text}
...

Final:
\end{promptbox}

\subsection{Rationale Generation Prompt}
\label{sec:prompt_rationale_generation}
After answer selection, we generate a concise rationale conditioned on the selected answer. The selected answer is provided explicitly, and the model is asked to explain why that answer is supported.

\begin{promptbox}{Answer-conditioned rationale generation prompt}
Answer the medical multiple-choice question.
The final answer has already been selected.
Write a concise medical rationale that supports this selected answer.

Output exactly in this format:
Rationale: <2-3 concise medical sentences explaining the key evidence>

Question:
{question}

Options:
{label}. {option text}
...

Selected answer:
{selected_answer_label}. {selected_answer_text}
\end{promptbox}

\subsection{Candidate Evidence-Span Proposal Prompt}
\label{sec:prompt_evidence_span}
We use Qwen2.5-72B-Instruct only as a high-recall candidate span proposer for cache construction. The model is used only as a candidate span proposer; proposed spans are subsequently filtered by the behavioral option-scoring tests described in Section~\ref{sec:faithfulness_cache} and Appendix~\ref{sec:cache_algorithm_appendix}.

\begin{promptbox}{Candidate evidence-span proposal prompt}
You are a key-evidence span extractor.

Given a question, options, selected answer, and rationale, extract minimal evidence spans.

Definition:
An evidence span is a short phrase or clause that supports the selected answer without directly naming the selected answer.

Rules:
1. Extract up to {MAX_EVIDENCE_UNITS} candidate evidence spans.
2. Each evidence span must be an exact substring of the rationale.
3. Do NOT include the selected answer option text: "{SELECTED_ANSWER_TEXT}".
4. Do NOT include the option label: "{SELECTED_ANSWER_LABEL}".
5. Do NOT include phrases such as "the correct answer is", "therefore", "thus", "the most appropriate choice", or "the selected answer".
6. If a sentence contains the selected answer term, extract only the clue portion that does not name it.
7. Prefer clinical/statistical/mechanistic clues, contrastive facts, symptoms, signs, lab or imaging findings, and causal links.
8. Avoid generic conclusions and meta-text such as "End of response".
9. Each span should usually be 3-15 words.
10. If a candidate is longer than {MAX_EVIDENCE_WORDS} words, split it into smaller evidence spans.
11. Return JSON only, exactly with this schema:
{"evidence_units":[{"text":"exact substring"}]}

Question:
{QUESTION}

Options:
{label}. {option text}
...

Selected answer:
{SELECTED_ANSWER_LABEL}. {SELECTED_ANSWER_TEXT}

Rationale:
{RATIONALE}
\end{promptbox}

\subsection{Recovery Prompt for PredRec and GoldRec}
\label{sec:prompt_recovery}
For predicted-answer reconstruction and gold-answer reconstruction, the recovery model receives the question, answer options, and generated rationale. We use deterministic candidate scoring rather than free-form generation. Each candidate option is appended after \texttt{Final:}, and the option with the highest length-normalized conditional log-likelihood is selected.

\begin{promptbox}{Rationale-based recovery prompt}
Infer the answer to the following medical multiple-choice question based only on the provided explanation.
Choose exactly one option.

Return the option label followed by the option text.

Question type: {question_type}

Question:
{question}

Options:
{label}. {option text}
...

Explanation:
{generated_rationale}

Final:
\end{promptbox}

\subsection{Full-precision Evidence Retention Claim Extraction Prompt}
\label{sec:prompt_fer_claim_extraction}
For full-precision Evidence Retention (FER), we first decompose the full-precision rationale into atomic clinical claims. 

\begin{promptbox}{FER atomic clinical-claim extraction prompt}
You are an expert medical evaluator for clinical multiple-choice QA.

Extract atomic clinical evidence claims from the full-precision rationale.

Each claim must contain exactly one clinical assertion that is relevant to supporting the selected answer.

Include clinical findings, lab findings, diagnoses, treatments, risk factors, pathophysiology, or other discriminative clinical evidence.

Do not include direct answer statements, option letters, or phrases such as "the answer is".

Do not add information that is not stated or clearly implied by the rationale.

Return only a JSON object with this schema:
{
  "claims": [
    {
      "claim_id": "c1",
      "claim": "<one atomic clinical claim>",
      "evidence_type": "<clinical_finding|lab_finding|diagnosis|treatment
      |risk_factor|pathophysiology|other>"
    }
  ],
  "reason": "<brief reason>"
}

Question:
{question}

Options:
{label}. {option text}
...

Full-precision selected answer:
{full_precision_selected_answer}

Full-precision rationale:
{full_precision_rationale}
\end{promptbox}

\subsection{Full-precision Evidence Retention Claim-Support Judge Prompt}
\label{sec:prompt_fer_support}
Given atomic clinical claims extracted from the full-precision rationale, the judge checks whether each claim is retained by a candidate rationale. A claim is counted as retained when the candidate rationale states, entails, or clearly preserves the clinical content of the full-precision claim. Contradicted claims are treated as not retained.

\begin{promptbox}{FER claim-support judgment prompt}
You are an expert medical evaluator for clinical multiple-choice QA.

Judge whether each full-precision clinical evidence claim is supported by the candidate rationale in the context of the same question and answer options.

Mark a claim as:
- supported: the candidate rationale states, entails, or clearly preserves the clinical content of the full-precision claim.
- unsupported: the candidate rationale does not contain enough information to support the claim.
- contradicted: the candidate rationale states something incompatible with the claim.

Do not judge whether the answer is correct.
Do not use outside medical knowledge except to interpret standard medical language.
Judge only whether the provided candidate rationale supports the full-precision claim in this question context.

Return JSON only, with one result for every claim_id:
{
  "claim_results": [
    {"claim_id": "c1", "support": "supported", "reason": "<short reason>"}
  ],
  "reason": "<brief overall note>"
}

Question:
{question}

Options:
{label}. {option text}
...

Full-precision selected answer:
{full_precision_selected_answer}

Full-precision rationale:
{full_precision_rationale}

Candidate system:
{candidate_system}

Candidate selected answer:
{candidate_selected_answer}

Candidate rationale:
{candidate_rationale}

Full-precision clinical evidence claims:
{claims}
\end{promptbox}

\subsection{Unsupported-Claim Rate Claim Extraction Prompt}
To compute UCR, we first decompose each system rationale into atomic clinical claims. Unlike FER, which extracts claim from the full-precision rationale and evaluates whether they are retained by a system rationale, UCR extracts claims from each system rationale and evaluates whether they are supported by the full-precision reference. 

\begin{promptbox}{UCR atomic clinical-claim extraction prompt}
You are an expert medical evaluator for clinical
multiple-choice QA. Follow the rubric exactly.
Output only the requested JSON object.

Extract atomic factual claims from the candidate rationale.

Definition:
- A claim is one medically or clinically meaningful factual
  statement, diagnostic assertion, mechanism, treatment
  statement, risk factor, or answer-supporting reasoning step.
- Extract claims only from the candidate rationale.
- Use the question and options only for disambiguation.
- Do not extract the answer letter itself as a claim.
- Exclude generic filler, uncertainty hedges, and purely
  stylistic text.
- Prefer short, atomic claims.
- Split compound statements when needed.
- Return at most 20 claims.
- Set truncated=true if additional factual claims were omitted
  because of this cap.
- If there is no factual medical claim, return an empty list
  and truncated=false.

Question type:
{question_type}

Question:
{question}

Options:
{label}. {option text}
...

Candidate selected answer:
{candidate_selected_answer}

Candidate rationale:
{candidate_rationale}

Return JSON only:
{
  "claims": [
    {
      "claim_id": "c1",
      "claim": "<one atomic claim>",
      "claim_type":
        "<clinical_finding|lab_finding|diagnosis|treatment|
        risk_factor|pathophysiology|general_medical_fact|
        answer_mapping|other>"
    }
  ],
  "reason": "<brief note on extraction>",
  "truncated": false
}
    
\end{promptbox}

\subsection{Unsupported-Claim Rate Claim-Support Judge Prompt}
Given atomic clinical claims extracted from each system rationale, the judge checks whether each claim is supported by the full-precision rationale. The question and answer options are provided only to disambiguate claim meaning. Claims labeled unsupported added or contradicted are counted as unsupported, while generally valid background claims are not penalized. Although the judge also provides a secondary lenient clinical judgment, only the primary teacher-source judgment is used to compute UCR.

\begin{promptbox}{UCR Claim-support Judgment prompt}
You are an expert medical evaluator for clinical
multiple-choice QA. Follow the rubric exactly.
Output only the requested JSON object.

Judge each candidate rationale claim under two reference
settings.

Primary teacher-source setting, used for the reported UCR:
- The active source is only the FP16 teacher rationale.
- The question and options are shown only to disambiguate
  claim meaning.
- Do not use the question or options to mark a claim as
  source_supported unless the same claim is stated or clearly
  entailed by the FP16 rationale.
- Do not use the gold answer, gold explanation, or standard
  medical knowledge to mark a claim as source_supported.
- The question is whether the quantized rationale introduced
  a new evidence claim that is not supported by the FP16
  teacher rationale.

Secondary lenient clinical setting:
- The active source is the FP16 teacher rationale plus clearly
  valid standard medical knowledge.
- The gold answer and gold explanation are intentionally
  excluded.
- This setting asks whether a new claim can still be clinically
  justified without using the gold explanation as a reference.

Labels for both settings:
- source_supported:
  The claim is stated or clearly entailed by the active source.

- valid_background:
  The claim is not stated in the active source, but is a
  generally valid medical background fact and does not
  introduce a new patient-specific finding, diagnosis,
  treatment effect, causal claim, or answer-specific evidence.

- unsupported_added:
  The claim introduces a new patient-specific condition,
  finding, diagnosis, treatment effect, causal or risk claim,
  or answer-supporting evidence that is not supported by the
  active source.

- contradicted:
  The claim contradicts the active source or standard medical
  knowledge.

- not_a_claim:
  The item is only an answer letter, formatting, generic filler,
  or is not a factual medical or clinical claim.

For fp16_overlap:
- present_in_fp16:
  The FP16 rationale states the same claim or a clear
  paraphrase.

- not_in_fp16:
  The claim is newly introduced relative to the FP16
  rationale.

- unclear:
  The overlap cannot be determined reliably.

Important:
Standard medical knowledge can make a claim valid_background
in the primary teacher-source setting, but it must not make the
claim source_supported unless it is also stated or clearly
entailed by the FP16 rationale.

Question type:
{question_type}

Question:
{question}

Options:
{label}. {option text}
...

FP16 teacher selected answer:
{full_precision_selected_answer}

FP16 teacher rationale:
{full_precision_rationale}

Candidate selected answer:
{candidate_selected_answer}

Candidate rationale:
{candidate_rationale}

Candidate claims to judge:
{candidate_claims}

Return JSON only, with one result for every claim_id:
{
  "claim_results": [
    {
      "claim_id": "c1",
      "fp16_overlap":
        "<present_in_fp16|not_in_fp16|unclear>",
      "teacher_support_status":
        "<source_supported|valid_background|unsupported_added|
        contradicted|not_a_claim>",
      "teacher_support_source":
        "<fp16_rationale|standard_medical_knowledge|none|
        not_applicable>",
      "clinical_support_status":
        "<source_supported|valid_background|unsupported_added|
        contradicted|not_a_claim>",
      "clinical_support_source":
        "<fp16_rationale|standard_medical_knowledge|none|
        not_applicable>",
      "reason": "<short reason>"
    }
  ],
  "reason": "<brief overall note>"
}
    
\end{promptbox}

\subsection{Pairwise Clinical-Rationale Judge Prompt}
\label{sec:prompt_pairwise_judge}
For pairwise judging, two systems are anonymized as Response A and Response B. A/B order is randomized outside the prompt, and the judge is given an explicit tie option. The comparison focuses on clinical correctness, support for the system's own selected answer, clinical sufficiency, and overall quality.

\begin{promptbox}{Pairwise clinical-rationale judgment prompt}
You are an expert medical evaluator.

Task:
Two anonymized model responses were written for the same medical multiple-choice question. Each response includes a predicted answer and an explanation. Choose which response is clinically better overall.

Compare them on these dimensions:
- clinical_correctness: Which response is more medically correct overall?
- answer_support: Which explanation better supports that response's own predicted answer?
- clinical_sufficiency: Which response provides more sufficient clinical evidence to justify its own predicted answer?

Comparison constraints:
- Compare the overall clinical quality of the two responses, not just style.
- Each response includes a predicted answer and an explanation.
- If the two responses predict the same answer, focus mainly on which explanation better justifies that shared answer.
- If the two responses predict different answers, compare which full response is clinically better overall, including whether the predicted answer itself is clinically more appropriate and whether the explanation justifies it.
- Judge answer_support and clinical_sufficiency relative to each response's own predicted answer.
- A wrong predicted answer can still be better supported by one explanation than the other, but overall it should usually lose to a clearly more clinically appropriate response.
- If an explanation only restates the predicted option without meaningful clinical reasoning or evidence, it should not beat a more reasoned explanation on answer_support or clinical_sufficiency.
- If an explanation is generic and could apply to multiple options, it should not beat a more specific explanation on clinical_sufficiency.
- If an explanation contains a central medical error, it should not beat a more medically correct explanation on clinical_correctness or clinical_sufficiency.
- If an explanation contradicts that response's predicted answer, it should not beat a supporting explanation on answer_support.
- If a response has no explanation, treat that as a weakness.
- Prefer Tie when the two explanations are effectively equal.
- Do not reward verbosity or confidence.

Question type:
{question_type}

Question:
{question}

Options:
{label}. {option text}
...

Response A predicted answer:
{response_A_predicted_answer}

Response A explanation:
{response_A_explanation}

Response B predicted answer:
{response_B_predicted_answer}

Response B explanation:
{response_B_explanation}

Output exactly one JSON object and nothing else, using only A, B, or Tie as values:
{"clinical_correctness": "A", "answer_support": "A", "clinical_sufficiency": "Tie", "overall": "A"}
\end{promptbox}

\subsection{Qualitative Analysis Prompt}
\label{sec:prompt_qualitative_analysis}
This analysis is illustrative and is not used as a quantitative metric. The two quantized systems are anonymized as System A and System B before being shown to the assistant, and the resulting summaries are manually inspected and edited by the authors.

\begin{promptbox}{Qualitative rationale-comparison prompt}
You are assisting a qualitative comparison of rationales generated by medical multiple-choice QA systems.

The reference system is the comparison target for preservation analysis, but do not assume it is clinically perfect.
The two quantized systems are anonymized as Quantized system A and Quantized system B.

Task:
Compare how each quantized system's rationale uses clinical evidence relative to the reference rationale.

Important rules:
- Return only the requested structured output.
- Do not write an overall comparison.
- Do not summarize or rewrite the generated rationales.
- Do not assign numerical scores.
- Do not decide which system is globally better.
- Focus only on evidence preservation, rationale drift, and clinically relevant differences.
- Keep each field concise.
- Ground the analysis in the provided question, options, predictions, and rationales.
- You may use general medical knowledge only to identify whether a rationale introduces an unsupported, irrelevant, or clinically questionable claim.
- If a claim is unsupported or less relevant, explain it briefly and tie it to the provided rationale or clinical item.

Input:

Question:
{question}

Options:
{options}

Gold answer:
{gold_answer}

Full-precision prediction:
{full_precision_prediction}

Full-precision rationale:
{full_precision_rationale}

Quantized system A prediction:
{system_A_prediction}

Quantized system A rationale:
{system_A_rationale}

Quantized system B prediction:
{system_B_prediction}

Quantized system B rationale:
{system_B_rationale}

Return exactly this JSON object:
{
  "system_A": {
    "preserved_evidence": ["<concise item>", "..."],
    "omitted_or_weakened_evidence": ["<concise item>", "..."],
    "unsupported_or_questionable_claims": ["<concise item>", "..."],
    "clinical_rationale_drift": "<one concise sentence>"
  },
  "system_B": {
    "preserved_evidence": ["<concise item>", "..."],
    "omitted_or_weakened_evidence": ["<concise item>", "..."],
    "unsupported_or_questionable_claims": ["<concise item>", "..."],
    "clinical_rationale_drift": "<one concise sentence>"
  },
  "brief_comparison": "<two concise sentences comparing evidence preservation only>"
}
\end{promptbox}

\section{Human Validation of FER Judgments}
\label{app:human_validation}

\paragraph{Sample and task}
We conduct a human validation on 20 Llama-3-8B-Instruct MedExpQA cases from the same-prediction subset used for FER. The 20 full-precision rationales contain 126 automatically extracted atomic clinical claims. Each claim is evaluated against two anonymized candidate rationales, resulting in 252 candidate--claim judgments per annotator. The candidate rationales correspond to OSTQuant and our method, but their identities are hidden during annotation.

\paragraph{Annotation protocol}
Two annotators complete the evaluation independently. For each case, annotators are shown the medical multiple-choice question, answer options, selected answer, full-precision reference rationale, two anonymized candidate rationales, and the extracted reference claims. For each claim, they judge whether each candidate rationale preserves the same clinical meaning. The claims are fixed in advance, and annotators do not evaluate the automatic claim-extraction stage. The examples are de-identified benchmark medical QA items, and annotations are reported only in aggregate.

\paragraph{Instructions given to annotators}
Annotators received the following instructions:

\begin{quote}
\small
The goal of this evaluation is to judge whether each candidate
explanation preserves the clinical meaning of the listed
reference claim.

For each case, you will see a medical multiple-choice question,
answer options, the selected answer, a full-precision reference
rationale, two anonymized candidate rationales, and a list of
atomic clinical claims extracted from the reference rationale.

For each listed claim, mark whether Candidate A and Candidate B
preserve the same clinical meaning as the reference claim.

Use the following labels:
\begin{itemize}[noitemsep,topsep=0pt,leftmargin=*]
    \item \textit{Supported}: the candidate rationale explicitly
    states the claim, clearly entails it, or provides an
    unambiguous paraphrase with the same clinical meaning.
    \item \textit{Not supported}: the claim is absent, too vague,
    only partially preserved, or contradicted by the candidate
    rationale.
\end{itemize}

Judge semantic preservation rather than exact wording. Do not
judge whether the selected answer is correct. Do not use this
task to evaluate whether the automatically extracted claims are
good or complete. The claims are fixed in advance. Focus only
on whether each candidate rationale supports each listed
reference claim.
\end{quote}

\paragraph{Annotator recruitment and compensation}
Two researchers from the authors' research group assisted with this small validation study after being invited by the authors. No external crowdsourcing platform was used, and no monetary compensation was provided. We acknowledge their assistance in the Acknowledgements.

\paragraph{Result analysis}
Both annotators completed all 252 candidate--claim judgments, with no missing labels. After de-anonymization, Candidate A corresponds to OSTQuant and Candidate B corresponds to our method. We compute agreement at the binary candidate--claim level, treating each judgment as \textit{Supported} or \textit{Not Supported}. For the GPT comparison, we compare each human label with the corresponding GPT-based label for the same candidate--claim pair.

\begin{table}[H]
\centering
\footnotesize
\setlength{\tabcolsep}{3.5pt}
\renewcommand{\arraystretch}{1.03}
\begin{tabular}{@{}lccc@{}}
\toprule
Comparison & Labels & Agree. & $\kappa$ \\
\midrule
Human--Human & 252 & 79.37 & 0.58 \\
GPT vs. Annotator 1 & 252 & 83.73 & 0.68 \\
GPT vs. Annotator 2 & 252 & 78.17 & 0.58 \\
GPT vs. pooled human & 504 & 80.95 & 0.63 \\
GPT vs. human consensus & 200 & 89.00 & 0.78 \\
\bottomrule
\end{tabular}
\caption{Human--human and human--GPT agreement for FER
judgments on the 20-case Llama-3/MedExpQA sample. Agreement
is computed over binary candidate--claim support labels.}
\label{tab:human_gpt_agreement}
\end{table}

The two annotators agree on 79.37\% of the binary support judgments, with Cohen's $\kappa=0.58$, indicating moderate agreement. Across all human--GPT label comparisons, the GPT-based judge agrees with the human labels in 80.95\% of cases, with Cohen's $\kappa=0.63$. On the subset where both human annotators agree, GPT agrees with the human consensus in 89.00\% of cases, with Cohen's $\kappa=0.78$. This suggests that the automatic judge is reasonably aligned with human annotations, especially on less ambiguous judgments.

\begin{table}[H]
\centering
\footnotesize
\setlength{\tabcolsep}{4pt}
\renewcommand{\arraystretch}{1.03}
\begin{tabular}{@{}lccc@{}}
\toprule
Evaluator & OSTQuant & Ours & $\Delta$ \\
\midrule
Annotator 1 & 47.62 & 57.14 & $+9.52$ \\
Annotator 2 & 57.94 & 67.46 & $+9.52$ \\
\midrule
Pooled human & 52.78 & 62.30 & $+9.52$ \\
GPT-based FER & 38.10 & 50.00 & $+11.90$ \\
\bottomrule
\end{tabular}
\caption{Human validation of FER on the 20-case
Llama-3/MedExpQA sample. Values are micro FER percentages.
$\Delta$ denotes Ours minus OSTQuant.}
\label{tab:human_validation_fer}
\end{table}

The human judgments show the same directional trend as the automatic FER judge. Pooling both annotators gives 52.78\% micro FER for OSTQuant and 62.30\% for our method, a +9.52 point difference. Thus, while the automatic judge is stricter in absolute support rate, both human and automatic evaluations agree directionally that our method better preserves full-precision clinical evidence on this sample.

\section{Pairwise Rationale Judging Details}
\label{app:pairwise_details}

Table~\ref{tab:pairwise_judge_medexpqa} reports the full
pairwise preference counts on MedExpQA same-answer cases.
The pairwise prompt and rubric are provided in
Appendix~\ref{sec:prompt_pairwise_judge}.

\begin{table}[t]
\centering
\footnotesize
\setlength{\tabcolsep}{3.5pt}
\renewcommand{\arraystretch}{1.02}
\begin{tabularx}{\columnwidth}{@{}Xccccc@{}}
\toprule
Model & Same pred. & OST win & Ours win & Tie & Ours\% \\
\midrule
OpenBioLLM & 44/125 & 9 & 18 & 17 & 66.67 \\
Llama-3 & 38/125 & 12 & 16 & 10 & 57.14 \\
BioMistral & 64/125 & 25 & 26 & 13 & 50.98 \\
Mistral & 71/125 & 17 & 29 & 25 & 63.04 \\
\midrule
Avg./pooled & 217/500 & 63 & 89 & 65 & 58.55 \\
\bottomrule
\end{tabularx}
\caption{Pairwise rationale preference counts on MedExpQA.
Ours\% excludes ties.}
\label{tab:pairwise_judge_medexpqa}
\end{table}

\section{Qualitative Examples}
\label{sec:appendix_qualitative}
We provide two MedExpQA examples from Llama-3-8B-Instruct to examine rationale drift when the final answer is preserved. In both experiments, the full-precision model, OSTQuant, and our method select the same option, allowing comparison of rationale content independent of answer differences.
We used GPT-5.4 API to assist a structured qualitative comparison of generated rationales. The quantized systems were anonymized during comparison, and the resulting summaries were manually inspected and edited by the authors. Table~\ref{tab:qualitative_main_cases} presents the comparisons, and the qualitative-analysis prompt is provided in Appendix~\ref{sec:prompts}.

\paragraph{Example 1}
OSTQuant preserves the broad neuropathic-pain interpretation and the selected answer, but introduces an unsupported medication-causation claim involving morphine and lactulose. In contrast, our method more closely follows the full-precision evidence structure and uses the normal neurological examination as a cue against focal nerve compression. This example illustrates how answer-preserving quantization can still alter the clinical justification.

\paragraph{Example 2}
OSTQuant preserves the broad worsening-heart-failure interpretation, but introduces a clinically relevant medication-class error by describing enalapril as a diuretic, which blurs the roles of ACE-inhibitor uptitration and intravenous diuresis. In contrast, our method retains the main full-precision evidence, including orthopnea, jugular venous distension, edema, and intravenous furosemide for fluid overload, although it omits some mechanistic detail about enalapril and afterload reduction. This example shows that even when the selected answer is unchanged, quantization can alter clinically important justification details.

\begin{table*}[t]
\centering
\scriptsize
\setlength{\tabcolsep}{3.0pt}
\renewcommand{\arraystretch}{1.07}
\begin{tabularx}{\textwidth}{@{}p{0.13\textwidth}YYY@{}}
\toprule

\multicolumn{4}{@{}p{\textwidth}@{}}{
\textbf{Example 1: Neuropathic pain management.}
\textbf{Question:}
A 66-year-old patient with stage IV pancreatic adenocarcinoma is taking delayed-release morphine.
He develops paresthesias and occasional lancinating pain in the right lumbar and periumbilical area that does not subside with rescue morphine.
Neurological examination is normal.

\textbf{Options:}
\textbf{A}: CT and evaluate nerve compression;
\textbf{B}: neurolytic treatment;
\textbf{C}: decrease morphine;
\textbf{D}: administer amitriptyline or gabapentin, dexamethasone, and increase morphine.
\textbf{Gold: D.}
} \\

\midrule

Aspect & FP & OSTQuant & Ours \\

\midrule

Prediction
& D
& D
& D \\

\midrule

Generated rationale
& The patient's symptoms of paresthesias and lancinating pain in the right lumbar and periumbilical area, despite adequate morphine treatment, suggest the presence of neuropathic pain. Amitriptyline or gabapentin, dexamethasone, and increasing the dose of morphine are effective in managing neuropathic pain, which is a common complication of pancreatic cancer.
& The patient's symptoms are consistent with neuropathic pain, which is a common complication of pancreatic cancer. The patient's history of delayed release morphine and lactulose may contribute to the development of neuropathic pain. The selected answer addresses the patient's symptoms by adding adjuvant therapy with amitriptyline or gabapentin, dexamethasone, and increasing the dose of morphine, which is a combination of treatments commonly used to manage neuropathic pain.
& The patient's symptoms of paresthesias and lancinating pain in the right lumbar and periumbilical area are consistent with neuropathic pain, which is a common complication of pancreatic cancer. The patient's normal neurological examination suggests that the pain is not due to compression of a specific nerve. Therefore, the most appropriate management is to try to control the neuropathic pain with medications such as amitriptyline or gabapentin, and dexamethasone, while also increasing the dose of morphine. This approach is supported by evidence that these medications can be effective in managing neuropathic pain in patients with cancer. \\

\midrule

Clinical evidence comparison
& Uses the key evidence: paresthesias, lancinating pain, inadequate response to morphine, pancreatic cancer context, and option D therapies.
& Preserves the neuropathic-pain interpretation and option D support, but adds an unsupported claim that morphine and lactulose may contribute to neuropathic pain.
& Preserves the reference evidence and adds relevant use of the normal neurological examination to argue against focal nerve compression. \\

\midrule

Takeaway
& Full-precision reference reasoning, though it assumes treatment effectiveness broadly.
& Mostly aligned with the FP answer, but shows rationale drift from an unsupported medication-causation claim.
& Closely aligned with the FP rationale and adds a relevant discriminative cue without major drift. \\

\midrule

\multicolumn{4}{@{}p{\textwidth}@{}}{
\textbf{Example 2: Decompensated heart failure management.}
\textbf{Question:}
A 73-year-old woman is admitted with progressive dyspnea at rest, orthopnea, and 4 kg weight gain.
Physical examination shows BP 150/84 mmHg, HR 100/min, increased jugular venous pressure, bibasal crackles, and malleolar edema.
Usual treatment: enalapril 5 mg every 12 hours, furosemide 80 mg/day.

\textbf{Options:}
\textbf{A}: administer intravenous furosemide;
\textbf{B}: increase enalapril dose according to tolerance and administer intravenous furosemide;
\textbf{C}: start a beta-blocker;
\textbf{D}: add amlodipine.
\textbf{Gold: B.}
} \\

\midrule

Aspect & FP & OSTQuant & Ours \\

\midrule

Prediction
& B
& B
& B \\

\midrule

Generated rationale
& The patient presents with signs of heart failure, including orthopnea, weight gain, and jugular venous distension, indicating a need for increased diuretic therapy. Increasing the dose of enalapril, a converting enzyme inhibitor, can help reduce afterload and improve cardiac function, while adding intravenous furosemide can further enhance diuresis and alleviate symptoms.
& The patient's symptoms of progressive dyspnea, orthopnea, and weight gain suggest worsening heart failure. Increasing the dose of enalapril, a diuretic, and administering intravenous furosemide will help to further reduce fluid overload and alleviate symptoms. This approach is supported by the patient's current treatment regimen, which includes a diuretic, and the need to further reduce fluid accumulation.
& The patient is presenting with signs of decompensated heart failure, including orthopnea, jugular venous distension, and pedal edema, indicating a need for increased diuretic therapy. Increasing the dose of enalapril and administering intravenous furosemide will help to reduce fluid overload and alleviate symptoms. \\

\midrule

Clinical evidence comparison
& Uses the key heart-failure evidence: orthopnea, weight gain, jugular venous distension, and need for intravenous diuresis; also gives an ACE-inhibitor uptitration rationale.
& Preserves worsening heart failure and fluid-overload evidence, but incorrectly calls enalapril a diuretic, blurring ACE-inhibitor and diuretic roles.
& Closely preserves the reference evidence, including decompensated heart-failure signs, jugular venous distension, edema, and intravenous furosemide for diuresis, but omits the afterload/cardiac-function rationale for enalapril. \\

\midrule

Takeaway
& Preserves the core clinical basis for option B.
& Mostly preserves the evidence but introduces a clinically relevant medication-class error.
& Strong evidence preservation with minor loss of ACE-inhibitor mechanism detail. \\

\bottomrule
\end{tabularx}

\caption{
Representative qualitative examples from the all-answer-match category.
All systems predict the same answer, but OSTQuant introduces rationale drift, while our method better preserves the full-precision model's answer-supporting clinical evidence.
}
\label{tab:qualitative_main_cases}
\end{table*}

\end{document}